\pdfoutput=1
\documentclass[11pt]{article}

\usepackage[preprint]{acl}
\usepackage{times}
\usepackage{latexsym}
\usepackage[T1]{fontenc}
\usepackage[utf8]{inputenc}
\usepackage{microtype}
\usepackage{inconsolata}
\usepackage{graphicx}

\usepackage{booktabs}
\usepackage{amssymb}
\usepackage{amsmath}
\usepackage{algorithm}
\usepackage[noend]{algpseudocode}
\usepackage{multirow}
\usepackage{placeins}
\usepackage{xcolor}
\usepackage{tcolorbox}
\tcbuselibrary{breakable,skins}
\usepackage{enumitem}

\definecolor{darkblue}{rgb}{0, 0, 0.5}
\definecolor{appxbrown}{rgb}{0.70, 0.52, 0.10}
\definecolor{promptblue}{RGB}{84,145,205}
\definecolor{promptbg}{RGB}{235,240,248}
\definecolor{promptaccent}{RGB}{78,145,220}
\definecolor{questionframe}{RGB}{156,109,46}
\definecolor{questionbg}{RGB}{249,242,230}
\definecolor{dahsframe}{RGB}{58,132,103}
\definecolor{dahsbg}{RGB}{232,245,238}
\definecolor{nonalignedframe}{RGB}{170,89,116}
\definecolor{nonalignedbg}{RGB}{249,236,242}

\newcommand{\promptvar}[1]{\textcolor{promptaccent}{\{#1\}}}
\newcommand{\thetaprev}{\theta_{\mathrm{\char111\char108\char100}}}
\ExplSyntaxOn
\NewDocumentCommand{\model}{m}
  {
    \tl_set:Nn \l_tmpa_tl {#1}
    \tl_replace_all:Nnn \l_tmpa_tl {-} {-\allowbreak}
    \tl_replace_all:Nnn \l_tmpa_tl {/} {/\allowbreak}
    \texttt{\tl_use:N \l_tmpa_tl}
  }
\ExplSyntaxOff

\newtcolorbox{prompttemplatebox}[2][]{
  enhanced,
  breakable,
  width=\linewidth,
  colback=promptbg,
  colframe=promptblue,
  colbacktitle=promptblue,
  coltitle=white,
  title={#2},
  boxrule=0.9pt,
  arc=3mm,
  left=4mm,
  right=4mm,
  top=2mm,
  bottom=2mm,
  toptitle=1.5mm,
  bottomtitle=1.5mm,
  fonttitle=\bfseries\large,
  before upper={
    \setlist[itemize]{topsep=0.35em,itemsep=0.2em,parsep=0pt,partopsep=0pt}
    \setlist[enumerate]{topsep=0.35em,itemsep=0.2em,parsep=0pt,partopsep=0pt}
  },
  #1,
}

\newcounter{subfigure}[figure]

\makeatletter
\providecommand*{\p@subfigure}{}
\makeatother

\title{Mitigating Distribution Sharpening in Math RLVR via Distribution-Aligned Hint Synthesis and Backward Hint Annealing}

\author{Pei-Xi Xie \and Che-Yu Lin \and Cheng-Lin Yang \\
  CyCraft AI Lab, Taiwan \\
  \texttt{\{peixi.xie, jerry.lin, cl.yang\}@cycraft.com}}

\begin{document}

\maketitle

\begin{abstract}
In reinforcement learning with verifiable rewards (RLVR), challenging math questions whose sampled groups remain uniformly incorrect yield no informative update, letting easier questions dominate actor updates. This imbalance can contribute to distribution sharpening: low-$k$ accuracy improves while problem coverage contracts. Hint scaffolding can restore informative updates by revealing a prefix of a verified teacher solution. However, existing hint-based approaches still face two issues: teacher–student distribution mismatch and the need to reduce hint exposure to match no-hint evaluation. We address these issues through two components. Distribution-Aligned Hint Synthesis (DAHS) constructs verified teacher hints conditioned on student-style responses. Backward Hint Annealing (BHA) anneals hint exposure across length buckets as bucket accuracy improves. BHA also uses per-question hint dropout to keep no-hint rollouts throughout RL training. Under the DAPO training framework, our method improves both pass@1 and pass@2048 over DAPO on AIME24/25/26 for \model{Qwen3-1.7B-Base} and \model{Qwen3-4B-Base}, and improves pass@1 on five secondary math benchmarks (Olympiad, MATH500, Minerva Math, AMC23, and GSM8K). On \model{Llama-3.2-1B-Instruct}, the gains concentrate in the high-$k$ regime without recovering the base model's coverage. Overall, when verified teacher prefixes are available, DAHS+BHA mitigates DAPO-induced contraction of high-$k$ problem coverage in the evaluated challenging-math RLVR settings.

\end{abstract}

\section{Introduction}
Reinforcement Learning with Verifiable Rewards (RLVR) has shown strong performance on mathematical reasoning tasks. Methods such as GRPO and DAPO have become widely adopted and have had substantial empirical impact~\citep{shao2024deepseekmath,guo2025deepseek,yu2025dapo}. However, in RLVR with group-relative policy updates, challenging questions may fail to generate informative updates for long periods when sampled groups remain uniformly incorrect. This optimization imbalance allows easier questions to dominate the update process. From a training-dynamics perspective, this imbalance can concentrate updates on a narrow subset of already reachable trajectory regions, leaving much of the effective trajectory support under-optimized. This pattern is consistent with recent analyses of distribution sharpening, where training can improve low-$k$ accuracy while narrowing effective solution coverage and leaving high-$k$ coverage under-improved~\citep{yue2025does,he2025rewarding}. This raises a practical question: \textit{how can we broaden continuation contexts to diversify learning and restore informative updates on challenging questions early in training?}

Recent hint-scaffolding methods~\citep{xi2024traininglargelanguagemodels,zhang2025breadbranchedrolloutsexpert,zhang2025scaf} suggest one possible answer: during training, they provide the student with a hint prefix from a teacher solution or another demonstration. The policy is then trained to continue from that prefix. In this line of work, a hint broadly refers to auxiliary information provided to help the model answer a question. Throughout this paper, a \textit{hint} therefore denotes a prefix of a complete, verified teacher solution~\citep{zhang2025breadbranchedrolloutsexpert}. In our view, hint continuation goes beyond providing extra information; it changes the continuation contexts under which the policy learns during training. By providing prefixes of varying lengths, hint scaffolding exposes the policy to continuations from multiple points along the reasoning trace, making challenging questions more likely to provide informative learning signals early in training~\citep{zhang2025breadbranchedrolloutsexpert}. In this sense, hint continuation can both restore early learning signals and reduce reliance on a narrow subset of already reachable modes.

Hint scaffolding also reshapes which questions can contribute training signal at all. When a question's sampled groups remain uniformly incorrect, group-relative updates provide no informative signal, and the set of learnable questions is therefore in practice bounded by what the student can already reach. Verified teacher scaffolds can move a subset of these questions into the learnable range: conditioning on a verified prefix raises the chance of a verified-correct continuation, extending the practical coverage ceiling from student-only reachability to teacher-scaffold reachability. We present this as an empirical coverage expansion, not a guarantee that the teacher-scaffold-reachable set strictly contains the student-reachable set.

However, prepending teacher hints does not directly achieve these effects~\citep{wang2025hinthelpingineffectiverollouts}. In our suffix-only setting, existing hint-based approaches still face two issues central to no-hint reasoning: a context mismatch between teacher hints and the student policy, and the gap between training with hints and no-hint evaluation.

\begin{figure*}[t]
\begin{center}
\includegraphics[width=0.9\textwidth]{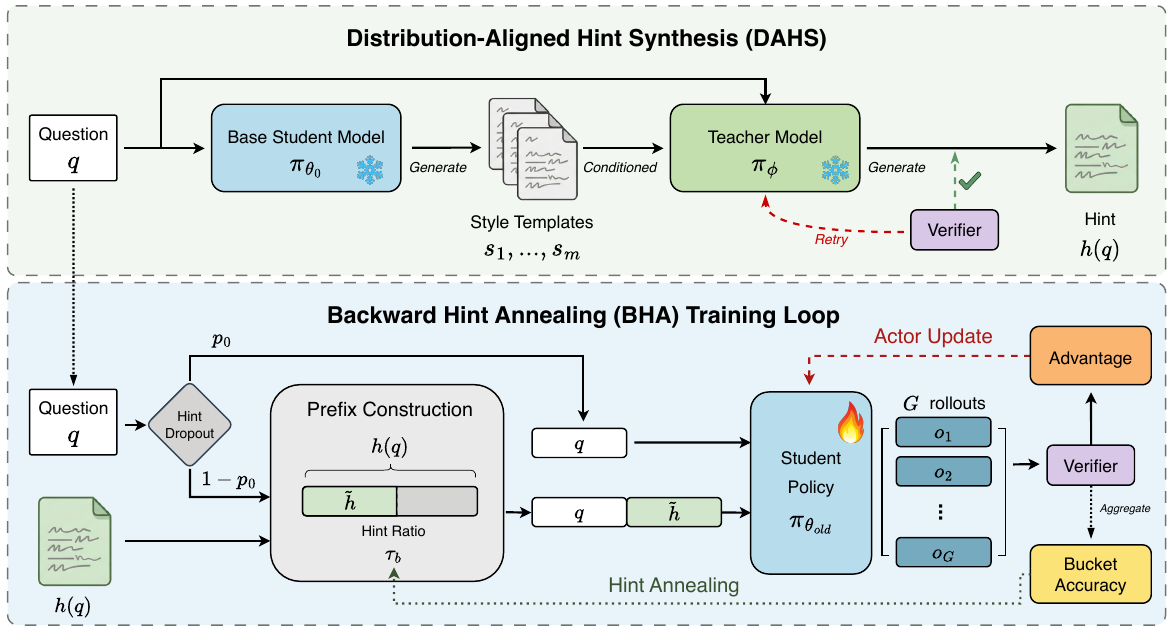}
\end{center}
\caption{\textbf{Overview of the framework.} \textbf{(Top)} Given a question $q$, the base student model generates style templates for the teacher model. Conditioned on both $q$ and these templates, the teacher generates a solution, and DAHS keeps it as the teacher hint if it passes the verifier. \textbf{(Bottom)} During RL training, hint dropout operates at the question level: the rollouts for a given question either receive no hint or share the same hint prefix derived from the teacher hint. When BHA applies hints, prefix construction truncates the full teacher hint $h(q)$ into the hint prefix $\tilde{h}$ according to the bucket-specific hint ratio $\tau_b$. The verifier outputs then serve two roles: they define the actor advantage and provide the bucket accuracy used to anneal hints. See Sec.~\ref{sec:dahs} for DAHS and Sec.~\ref{sec:bha} for BHA.}
\label{fig:overview}
\end{figure*}

\begin{figure}[t]
\begin{center}
\includegraphics[width=0.95\linewidth]{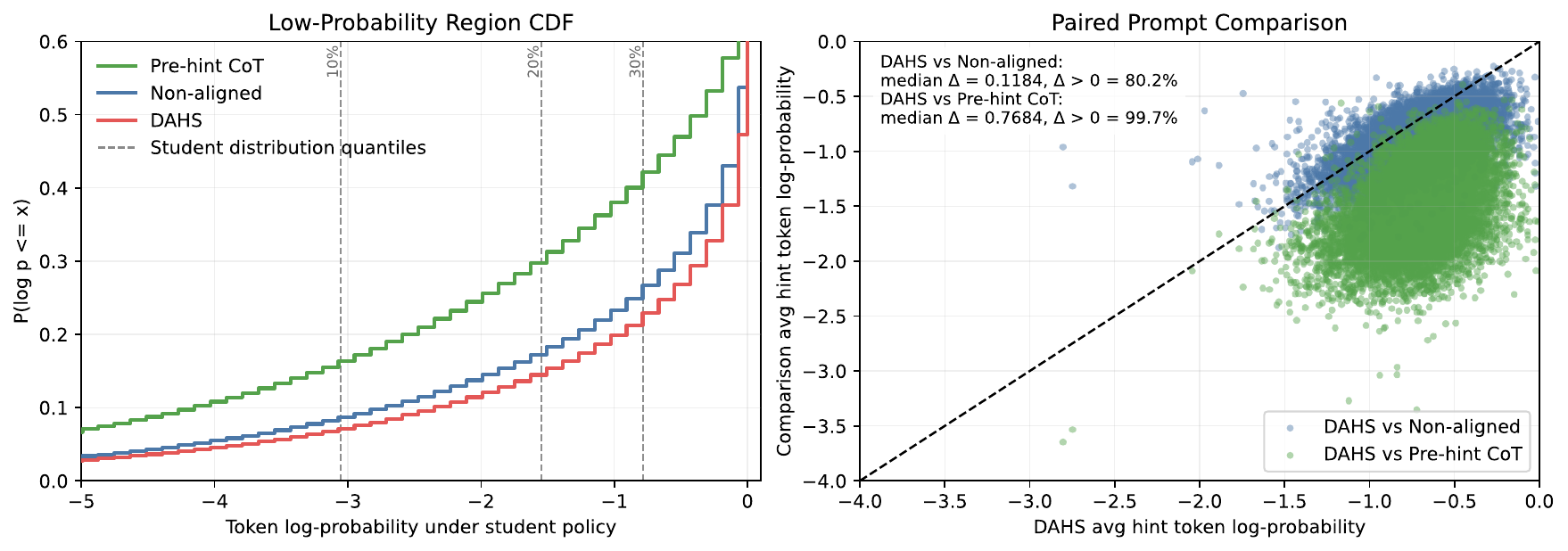}
\end{center}
\caption{\textbf{Hint log-probability comparison under the student policy on \model{Qwen3-1.7B-Base}~\citep{qwen3technicalreport}, with the three compared data sources generated by \model{gpt-oss-120b}~\citep{openai2025gptoss120bgptoss20bmodel}.} Pre-hint CoT is the teacher model's original chain-of-thought before hint synthesis. Non-aligned hints employ verified teacher-generated hint segments without student-style conditioning. DAHS hints employ verified teacher-generated hint segments with student-style conditioning.}
\label{fig:dahs-logprob-compare}
\end{figure}

\textbf{1) Context mismatch between teacher hints and the student policy}. Tokens in teacher hints, including detailed chain-of-thought when present, can have low probability under the student policy, and the mismatch grows when the teacher and student come from separate model families, as illustrated in Fig.~\ref{fig:dahs-logprob-compare}. This figure serves as the motivation for our hint-synthesis design; in practice, better-aligned hints indeed prove effective, as shown by the DAHS ablation in Sec.~\ref{sec:ablation-dahs}. Even under suffix-only updates, directly prepending such hints can shift the hint prefix into the student's low-probability regions and place the continuation in an unnatural or out-of-distribution context. This can yield brittle continuations and weaken transfer to no-hint evaluation.

\textbf{2) The gap between training with hints and no-hint evaluation}. Under suffix-only on-policy updates, hints no longer teach the policy to reproduce the teacher. Instead, they change the continuation-context distribution seen by the policy during training, affecting both whether a question yields informative updates and whether the policy later over-relies on the hint scaffold. The hint scaffold therefore becomes a central design variable: how much hint to reveal, how to retain no-hint rollouts, and how to anneal hints over training.

To address these two issues, we propose a teacher-hint synthesis method, \textit{Distribution-Aligned Hint Synthesis} (DAHS), which addresses the context mismatch that arises under suffix-only training with hints, and a complementary training method, \textit{Backward Hint Annealing} (BHA), which addresses the transfer gap between training with hints and no-hint evaluation. Under verified teacher-prefix availability and no-hint evaluation, DAHS+BHA mitigates the DAPO-induced contraction of high-$k$ problem coverage that we observe on competition-level math (AIME24/25/26)~\citep{aime24,aime25,aime26}. We measure this contraction, relative to the base model, as a drop in the number of questions solved at least once or in pass@high-$k$ (Sec.~\ref{sec:main-results}). This effect holds on both \model{Qwen3-1.7B-Base} and the larger \model{Qwen3-4B-Base}.

\textbf{Our main contributions and findings are as follows:}
\begin{enumerate}
    \item We characterize a failure mode in math RLVR that can contribute to distribution sharpening: challenging questions can remain without informative updates for long periods, letting easier questions dominate actor updates.
    \item We propose a hint-scaffolding framework that combines DAHS, which synthesizes verified teacher hints aligned with the student policy, with BHA, which broadens continuation contexts during training and gradually anneals hint exposure back to no-hint evaluation.
    \item Our ablations show that better-aligned hints correlate with stronger high-$k$ gains, that gradual hint reduction with per-question hint dropout improves no-hint transfer, and that length-bucketed annealing avoids the failure mode of a global schedule while offering a more practical trade-off than per-prompt hint search.
\end{enumerate}

\section{Preliminaries}
\label{sec:preliminaries}
\subsection{Verifier-based reward}
We employ a verifier-based reward~\citep{shao2024deepseekmath,wen2025reinforcementlearningverifiablerewards} defined by a deterministic equivalence check between the extracted final answer and the ground-truth answer.
In Eq.~\ref{eq:rule-reward}, $q$ is the question, $a$ is the ground-truth answer, and $\hat{a}(o)$ is the final answer extracted from response $o$.
\begin{equation}
\label{eq:rule-reward}
	    R(q,o)
    \;=\;
	    \left\{
	    \begin{array}{ll}
	    1, & \mbox{if }\texttt{is\_equivalent}(\hat{a}(o), a),\\[3pt]
	    0, & \mbox{otherwise}
	    \end{array}
	    \right.
\end{equation}

\subsection{Dynamic sampling in DAPO}
\label{sec:dynamic-sampling}
Following DAPO~\citep{yu2025dapo}, a group becomes uninformative under group-relative advantage normalization if responses in the group receive the same reward, yielding zero within-group reward variance. 
This arises under verifier-based rewards for uniformly correct or uniformly incorrect groups~\citep{nan2025ngrponegativeenhancedgrouprelative}.
DAPO addresses this via group \emph{over-sampling} followed by \emph{filtering}.
For each question $q$, define the empirical group accuracy $\bar{z}=\frac{1}{G}\sum_{i=1}^G \mathbf{1}\{R(q,o_i)=1\}$.
Dynamic sampling discards groups with $\bar{z}=0$ or $\bar{z}=1$ and retains only mixed groups with $0<\bar{z}<1$.
The sampler repeats generation and filtering until it fills a training batch with retained groups, making non-degenerate group-relative advantages drive the retained updates. 
Dynamic sampling filters such degenerate groups, which stabilizes training but does not restore a learning signal for questions that remain uniformly incorrect.

\subsection{Backward algorithm}
\label{sec:backward-algorithm}
The backward algorithm of \citet{salimans2018learningmontezumasrevengesingle} follows an end-to-start curriculum: it begins by resetting the environment to configurations near success and gradually moves those starting points backward, thereby requiring the policy to complete increasingly longer trajectories to reach success.
In LLM reasoning, BREAD~\citep{zhang2025breadbranchedrolloutsexpert} adopts a related prefix-revealing idea at the question level: when a question's sampled group is uniformly incorrect, it reveals a prefix of a verified expert solution and stops at the first prefix length that yields non-zero group accuracy, without continuing the search for the shortest such prefix.
In our setting, we apply the backward-curriculum idea of \citet{salimans2018learningmontezumasrevengesingle}, rather than BREAD's stop-at-first-signal selection, to hint placement: instead of adding or removing hints at the question level~\citep{zhang2025breadbranchedrolloutsexpert,zhang2025scaf}, we start with a longer revealed prefix of a complete hint and then gradually reveal less of that hint, thereby shrinking the visible prefix and requiring the model to perform increasingly more of the reasoning on its own.

\section{Method}
\label{sec:method}

\subsection{Distribution-Aligned Hint Synthesis}
\label{sec:dahs}
DAHS synthesizes one verified teacher hint per question by conditioning the teacher on a \emph{set} of student-style responses. Define $\pi_{\theta_0}$ as the base student model and $\pi_\phi$ as the teacher model. For each question $q$ with ground-truth answer $a$, we first draw $m$ student responses $s_j \sim \pi_{\theta_0}(\cdot \mid q)$, $j=1,\dots,m$, and treat the collection $\mathcal{S}(q)=\{s_j\}_{j=1}^m$ as style templates for the teacher, capturing the student's current solution patterns regardless of correctness. We then provide the original question together with $\mathcal{S}(q)$ to the teacher and ask for a single teacher solution $h \sim \pi_\phi(\cdot \mid q, \mathcal{S}(q))$.
We retain $h$ only if it passes the verifier in Eq.~\ref{eq:rule-reward}, i.e., $R(q,h)=1$.
We exclude questions without a verified teacher hint from the filtered training set shared across methods (sampling details in Appendix Sec.~\ref{app:exp-details}). Otherwise, we cache the verified teacher hint as $h(q)$.
Fig.~\ref{fig:dahs-logprob-compare} shows that DAHS hint prefixes better match the student policy distribution than hint sources used in prior work.

\subsection{Backward Hint Annealing}
\label{sec:bha}
\begin{figure}[t]
\centering
\includegraphics[width=\linewidth]{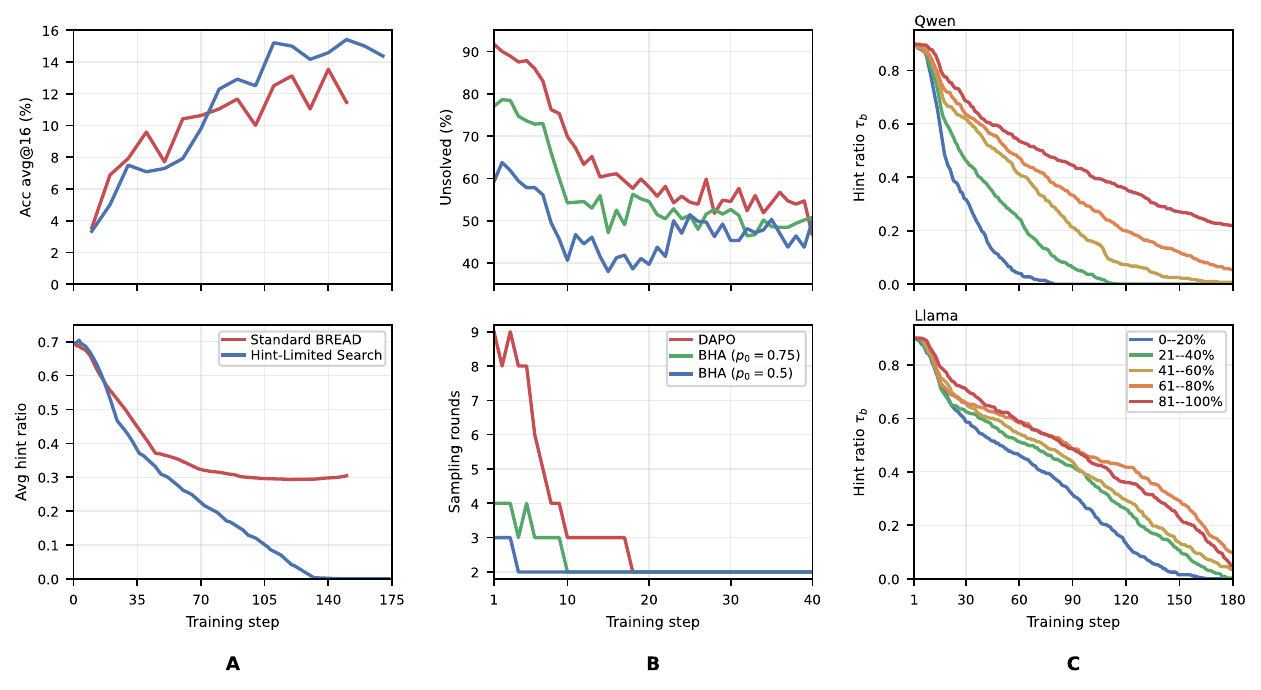}
\caption{\textbf{A:} no-hint transfer. \textbf{B:} early training dynamics over all questions, including no-hint ones under hint dropout. \textbf{C:} BHA bucketed hint-ratio annealing.}
\label{fig:hint-diagnostics}
\setcounter{subfigure}{0}
\refstepcounter{subfigure}\label{fig:no-dropout-eval-aime24}
\refstepcounter{subfigure}\label{fig:training-dynamics-dapo-bha}
\refstepcounter{subfigure}\label{fig:tau-group-bha075}
\end{figure}

BHA is a bucket-level backward curriculum that gradually reduces each bucket's hint ratio $\tau_b$ as bucket accuracy improves, enabling the transition from hint-scaffolded training to no-hint evaluation.
At evaluation time, the policy must solve questions without hints.
Fig.~\ref{fig:no-dropout-eval-aime24} shows why annealing matters: BREAD tends to retain longer hints, but explicit hint reduction yields better no-hint transfer.

\subsubsection{Hint construction}
\label{sec:hint-construct}
For each question $q$, we maintain a hint candidate pool $\mathcal{H}(q)$.
At initialization, $\mathcal{H}(q)$ contains the verified teacher hint $h(q)$.
We define the bucket index $b=b(q)\in\{1,\dots,B\}$ and employ $\tau_b$ to control how much of the selected hint we reveal.

When we employ a hint, we first draw a hint source $h^\star(q)\sim \mathcal{H}(q)$.
We then set $\tau=\tau_b$, compute a nominal cutoff $c_0=\mathrm{round}(\tau\,|h^\star(q)|)$, and draw a stochastic integer cutoff
\begin{equation}
\label{eq:cutoff-draw}
c \sim \operatorname{DiscreteUniform}\!\left(\left\{\max(0, c_0 - D), \dots, c_0\right\}\right).
\end{equation}
where the window size $D$ follows the local-window design of the backward algorithm~\citep{salimans2018learningmontezumasrevengesingle}.
We reveal the prefix $\tilde{h}=\mathrm{prefix}(h^\star(q),c)$ and prompt the student with $(q,\tilde{h})$ to generate an on-policy continuation.
We cap the generation length per prompt, with generated continuation length bounded by $L_{\max}-|\tilde{h}|$.

\subsubsection{Per-question hint dropout}
To keep the no-hint rollouts present throughout training, we apply \emph{per-question hint dropout}.
We provide an empty prefix with probability $p_0$; dynamic sampling subsequently determines whether the resulting group enters the actor update.
During such no-hint rollouts, if the policy model finds a verified correct solution for question $q$, we add the resulting solution trace to $\mathcal{H}(q)$.
This lets future updates with hints draw from both the teacher-provided hint and successful trajectories previously discovered by the policy itself, in the spirit of Go-Explore~\citep{ecoffet2021goexplorenewapproachhardexploration}.
The pool $\mathcal{H}(q)$ is thus refreshed during training, and the scaffold increasingly contains trajectories from the current policy rather than a fixed teacher style.

\subsubsection{Length-bucketed hint annealing}
\label{sec:length-bucket}
Per-question searching of $\tau$ can have high variance because each question provides only a limited number $G$ of sampled answers for estimating $\tau$.
Another drawback is that it can consume substantial compute on questions that remain uniformly incorrect.
We instead anneal hint ratios at the bucket level.
Our default design uses length-bucketed $\tau_b$.
Teacher-hint length serves as a proxy for question difficulty. We therefore partition the training set into $B$ length buckets.
In our data, longer teacher hints correlate with harder questions, and this proxy needs neither manual annotation nor an extra LLM-based difficulty classifier.
Ratio-based truncation remains comparable within each bucket because questions of similar length lose a similar number of hint tokens at each annealing step.
Fig.~\ref{fig:tau-group-bha075} shows the length buckets grouped into five 20\% bands.
Lower-length bands anneal earlier, whereas higher-length bands anneal later.
This supports bucket-wise annealing because each band can reduce hint exposure according to its own difficulty level.

For each question identifier $\mathrm{uid}$, we draw a group of $G$ on-policy answers and score each answer with the verifier.
We employ a binary correctness indicator $z_{\mathrm{uid},i}\in\{0,1\}$ for annealing and for dynamic sampling.
Using $\mathcal{U}_b$ to denote the set of question identifiers in bucket $b$ that receive a non-empty prefix, we compute the prompt-level and bucket-level accuracies as
\begin{equation}
\label{eq:prompt-acc}
    \bar{z}_{\mathrm{uid}} = \frac{1}{G}\sum_{i=1}^{G} z_{\mathrm{uid},i}, \quad \bar{z}_b = \mathrm{mean}_{\mathrm{uid}\in \mathcal{U}_b} \bar{z}_{\mathrm{uid}}.
\end{equation}
We compute $\bar{z}_b$ only when $\mathcal{U}_b\neq\emptyset$; otherwise, we leave bucket $b$ unchanged for that attempt.
When $\bar{z}_b \ge \eta$, we reduce the hint ratio by a fixed step size:
\begin{equation}
\label{eq:bucket-anneal}
    \tau_b \;\leftarrow\; \max(\tau_b-\Delta\tau,\,0),
\end{equation}
where $\eta$ is the annealing threshold and $\Delta\tau$ is a step size.
In the main \model{Qwen3-1.7B-Base} run, this schedule anneals the mean hint ratio from $0.90$ to $0.056$; among rollouts that receive a prefix, the realized visible fraction of the selected hint averages $0.29$ across training and $0.061$ over the final 10\% of steps (Appendix Sec.~\ref{app:default-hparams}).

\subsection{Reinforcement Learning Training Loop}
\label{sec:training-alg}
We apply dynamic sampling, as described in Sec.~\ref{sec:dynamic-sampling}, to retain only mixed groups for the actor update.
Mixed-group filtering limits the influence of flawed scaffolds: final-answer verification does not guarantee correct intermediate reasoning, but a corruption probe (Appendix Sec.~\ref{app:corrupted-hint}) shows that a flawed reasoning step sharply lowers the rate of verified-correct continuations, so such prefixes tend to yield uniformly incorrect groups that dynamic sampling filters.
Appendix Sec.~\ref{app:error-copying} audits the frequency of flawed-but-verified teacher solutions in practice.
Appendix Algorithm~\ref{alg:training-alg} summarizes the full procedure.

\subsubsection{Suffix-only on-policy policy gradient}
\label{sec:suffix-only}
Teacher hint tokens act only as context and do not receive gradients.
We update the policy only on tokens generated by the student~\citep{nath2025adaptiveguidanceacceleratesreinforcement,zhang2025breadbranchedrolloutsexpert}, which avoids mixing SFT-style losses into the hint prefix and avoids heuristic off-policy weighting or correction.
The trained object is thus the evolving policy's continuation behavior, not imitation of a fixed hint; Sec.~\ref{sec:main-results} tests this by comparing continuations from $h(q)$ and from alternative verified solutions after training.

For question $q$, we construct the drawn prefix $\tilde{h}$ by following the hint-construction and per-question hint-dropout procedures described earlier in this section.
We then draw a group of $G$ on-policy continuations $\{o_i\}_{i=1}^G$ from the training policy conditioned on $(q,\tilde{h})$.
We compute outcome rewards $R_i := R(q,o_i)$ with the same verifier as in the no-hint setting.
We omit the KL penalty in our main setup, following recent KL-free RLVR settings~\citep{yu2025dapo,zhang2025scaf,yan2025learningreasonoffpolicyguidance}, and apply the DAPO objective to the suffix tokens:

\begin{equation}
\label{eq:bha-obj}
    \begin{aligned}[b]
        &\mathcal{J}_{\mathrm{BHA}}(\theta)
        =
        \mathbb{E}_{\substack{(q,a)\sim \mathcal{D},\textcolor{blue}{\tilde{h}\sim \mathcal{P}_{\mathrm{BHA}}(\cdot\mid q)},\\ \{o_i\}_{i=1}^G\sim\pi_{\thetaprev}(\cdot\mid q, \textcolor{blue}{\tilde{h}})}}
        \bigg[
        \frac{1}{G}\sum_{i=1}^G
        \frac{1}{|o_i|}\sum_{t=1}^{|o_i|}
        \\[-2pt]
        &\min(r_{i,t}(\theta)\widehat{A}_i,\!\mathrm{clip}(r_{i,t}(\theta),1{-}\epsilon_{\mathrm{low}},1{+}\epsilon_{\mathrm{high}})\widehat{A}_i)
        \bigg]
    \end{aligned}
\end{equation}

where $\mathcal{P}_{\mathrm{BHA}}(\cdot\mid q)$ is the prefix-sampling distribution induced by the hint construction and per-question hint dropout described earlier in this section.
We compute the token-level importance ratio conditioned on the question, realized prefix, and prior generated tokens, and normalize the group-relative advantage within each sampled group:
\begin{equation}
\label{eq:bha-ratio}
    \begin{aligned}
    r_{i,t}(\theta)
    &=
    \frac{\pi_\theta(o_{i,t}\mid q, \textcolor{blue}{\tilde{h}}, o_{i,<t})}{\pi_{\thetaprev}(o_{i,t}\mid q, \textcolor{blue}{\tilde{h}}, o_{i,<t})}
    ,\\
    \widehat{A}_i
    &=
    \frac{
    R_i - \mathrm{mean}\!\left(\{R_j\}_{j=1}^G\right)
    }{
    \mathrm{std}\!\left(\{R_j\}_{j=1}^G\right)
    }
    \end{aligned}
\end{equation}

\section{Experiments}
\label{sec:experiments}

\subsection{Baseline}
\label{sec:baseline}
Across comparisons, we employ the same number of RL training steps and the same rule-based verifier; unless noted otherwise, other settings follow DAPO~\citep{yu2025dapo}.
\begin{itemize}[nosep,leftmargin=*]
    \item \textbf{DAPO.} Standard DAPO with dynamic sampling~\citep{yu2025dapo}, without hint scaffolding.
    \item \textbf{SFT (forward distillation).} Supervised fine-tuning on DAHS hints only, without RL.
    \item \textbf{BREAD.} BREAD~\citep{zhang2025breadbranchedrolloutsexpert} with dynamic sampling and DAHS hints.
    \item \textbf{Hint-Limited Search.} A compute-heavy BREAD variant with DAHS hints that binary-searches, per prompt, for the smallest non-degenerate hint ratio under a decaying global limit (Appendix Sec.~\ref{sec:ablation-perprompt-search}).
\end{itemize}

\subsection{Experimental Setup}
We evaluate DAHS+BHA under the DAPO training framework with verifiable rewards.
Our training pipeline builds on the \texttt{verl} framework~\citep{sheng2024hybridflow}.
These experiments focus on mathematics: training uses the DAPO-Math-17k dataset~\citep{yu2025dapo}, deduplicated by question, and evaluation uses AIME24/25/26~\citep{aime24,aime25,aime26} in the no-hint setting.
We select these datasets to match the failure mode we study: DAPO-Math-17k aligns our training corpus with the DAPO baseline and leaves 39.7\% of its questions without informative updates under vanilla DAPO (Sec.~\ref{sec:main-results}), and AIME is difficult enough for high-$k$ coverage contraction to appear.
We additionally report results on Olympiad~\citep{he2024olympiadbench}, MATH500~\citep{hendrycks2021measuringmathematicalproblemsolving}, Minerva Math~\citep{lewkowycz2022solvingquantitativereasoningproblems}, AMC23, and GSM8K~\citep{cobbe2021gsm8k} as additional benchmarks.

We run experiments on three student models: \model{Qwen3-1.7B-Base}, \model{Qwen3-4B-Base}~\citep{qwen3technicalreport}, and \model{Llama-3.2-1B-Instruct}~\citep{grattafiori2024llama3herdmodels}.
Our primary focus in this line of work is RL on base models~\citep{guo2025deepseek,zeng2025simplerlzooinvestigatingtamingzero}; therefore, the Qwen3 base models serve as the main base-model testbed.
Because base checkpoints outside the Qwen family yielded insufficient verified-correct responses on DAPO-Math-17k in our preliminary runs (near-zero keepable rollouts for \model{Llama-3.2-1B} and \model{Llama-3.2-3B}, and \model{gemma-3-1b-pt}~\citep{gemmateam2025gemma3}; see Appendix Sec.~\ref{app:base-boundary}), we employ \model{Llama-3.2-1B-Instruct} instead for stable RLVR training.
We train each method on the same filtered subset of questions for which DAHS obtains a verified teacher hint.
On DAPO-Math-17k, DAHS with our default open-weight teacher \model{gpt-oss-120b}~\citep{openai2025gptoss120bgptoss20bmodel} obtains a verified hint for 16,449 of the 17,398 deduplicated questions (94.55\%) at a mean of 1.39 teacher calls per question; hint synthesis is a one-time offline preprocessing step, and online RL adds no extra rollouts over vanilla DAPO (Appendix Sec.~\ref{app:teacher-yield}).
Appendix~\ref{app:exp-details} and Appendix~\ref{app:passk-protocol} cover the additional implementation details and the pass@$k$ protocol; Table~\ref{tab:hparams} lists the default hyperparameters.

\begin{figure}[!t]
\centering
\includegraphics[width=\linewidth]{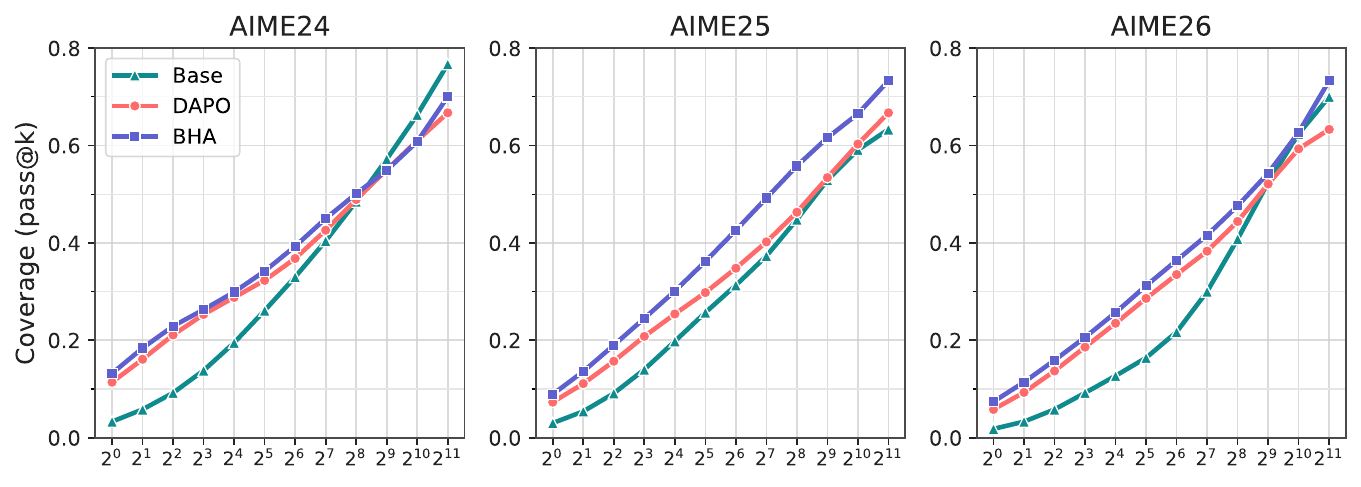}\\[2pt]
\includegraphics[width=\linewidth]{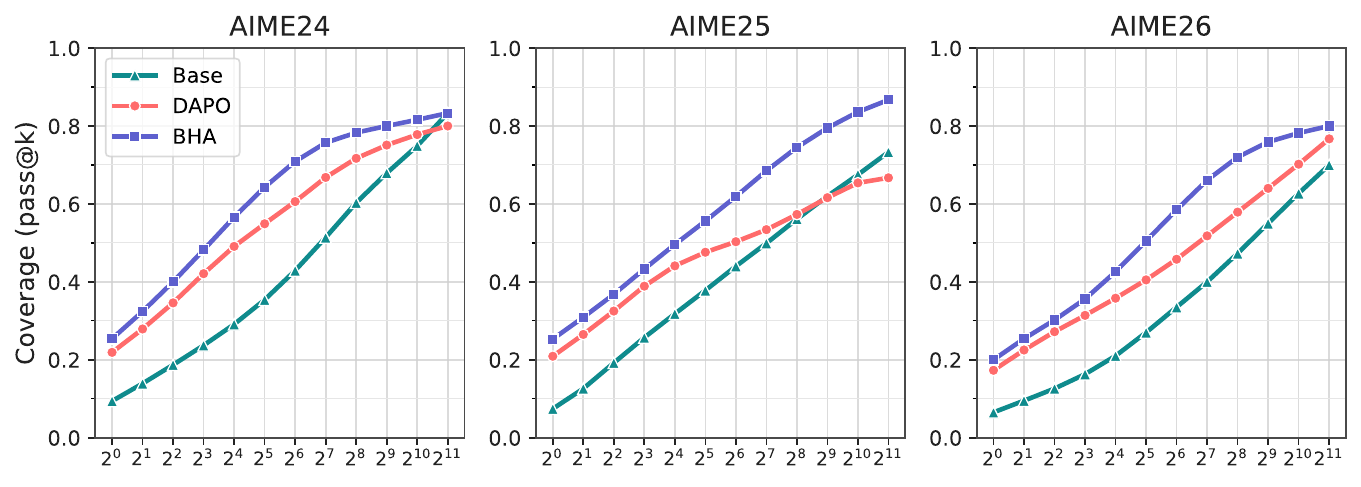}
\caption{Pass@$k$ curves on AIME24/25/26 for \model{Qwen3-1.7B-Base} (top) and \model{Qwen3-4B-Base} (bottom).}
\label{fig:passk-curves}
\end{figure}

\begin{table*}[!t]
\centering
\small
\setlength{\tabcolsep}{3pt}
\begin{tabular}{@{}ll*{6}{c}@{}}
\toprule
& & \multicolumn{2}{c}{\bf AIME24}
& \multicolumn{2}{c}{\bf AIME25}
& \multicolumn{2}{c}{\bf AIME26} \\
\cmidrule(lr){3-4}\cmidrule(lr){5-6}\cmidrule(lr){7-8}
{\bf Model} & {\bf Method}
& {\bf pass@1} & {\bf pass@2048}
& {\bf pass@1} & {\bf pass@2048}
& {\bf pass@1} & {\bf pass@2048} \\
\midrule

\multirow{6}{*}{Qwen3-1.7B}
& Base      & 3.3  & \textbf{76.7} & 3.0 & 63.3 & 1.8 & \underline{70.0} \\
& SFT       & 2.4  & \textbf{76.7} & 1.5 & \underline{70.0} & 2.3 & \underline{70.0} \\
& DAPO      & 11.4 & 66.7 & 7.3 & 66.7 & 5.8 & 63.3 \\
& BREAD  &   12.6   &   66.7   &   5.9  &   \underline{70.0}   &   6.7  &   66.7 \\
& HLS & \textbf{13.9} & 60.0 & \underline{7.8} & \textbf{73.3} & \underline{6.9} & 66.7 \\
& BHA (Ours)& \underline{13.2} & \underline{70.0} & \textbf{8.9} & \textbf{73.3} & \textbf{7.4} & \textbf{73.3} \\
\addlinespace[3pt]

\multirow{3}{*}{Qwen3-4B}
& Base      & 9.4  & \textbf{83.3} & 7.4 & \underline{73.3} & 6.5 & 70.0 \\
& DAPO      & \underline{21.9} & 80.0 & \underline{20.9} & 66.7 & \underline{17.3} & \underline{76.7} \\
& BHA (Ours)& \textbf{25.4} & \textbf{83.3} & \textbf{25.3} & \textbf{86.7} & \textbf{20.0} & \textbf{80.0} \\
\addlinespace[3pt]

\multirow{3}{*}{Llama3.2}
& Base      &   0.9   &   \textbf{70.0}   &  \textbf{0.1}   &   \textbf{56.7}   &  \underline{0.3}   &   \textbf{70.0} \\
& DAPO      &   \textbf{2.6}   &   43.3   &   0.0  &   23.3   &  \underline{0.3}   &   20.0 \\
& BHA (Ours)       &   \underline{2.4}   &   \underline{56.7}   &   \textbf{0.1}  &   \underline{26.7}   &  \textbf{0.6}   &   \underline{23.3} \\

\bottomrule
\end{tabular}
\caption{Results on AIME24/25/26. Full baseline comparisons are reported at 1.7B; the 4B block tests whether the coverage behavior of DAPO and DAHS+BHA persists at a larger Qwen scale. SFT, BREAD, Hint-Limited Search (HLS), and BHA (Ours) all use the same DAHS hints.}
\label{tab:aime-results}
\end{table*}

\subsection{Main Results}
\label{sec:main-results}

\textbf{DAHS+BHA improves no-hint performance across the pass@$k$ range.}~On \model{Qwen3-1.7B-Base} and \model{Qwen3-4B-Base}, DAHS+BHA outperforms DAPO at both pass@1 and pass@2048 on AIME24/25/26 (Table~\ref{tab:aime-results}; solved-at-least-once counts in Appendix Table~\ref{tab:solved-counts}). The pass@$k$ curves (Fig.~\ref{fig:passk-curves}) show gains over a broad range of $k$ at 1.7B and at every evaluated $k$ at 4B. At 4B, DAPO's pass@2048 falls below the base model on AIME24 and AIME25, whereas DAHS+BHA matches or exceeds it; the coverage contraction thus persists at the larger scale, and DAHS+BHA mitigates it. On \model{Llama-3.2-1B-Instruct}, pass@1 remains low, likely reflecting the limited capacity of the model; DAHS+BHA improves high-$k$ performance over DAPO but stays below the base model's coverage, and we treat this setting as a boundary case (Appendix Sec.~\ref{app:llama-passk}). On the secondary benchmarks, DAHS+BHA exceeds DAPO at pass@1 at both scales, but high-$k$ coverage is mixed (Appendix Sec.~\ref{app:secondary-benchmarks}).

\begin{table}[!t]
\centering
\small
\setlength{\tabcolsep}{4pt}
\begin{tabular}{@{}cccc@{}}
\toprule
& \multicolumn{2}{c}{\bf Qwen3-1.7B} & {\bf Qwen3-4B} \\
\cmidrule(lr){2-3}\cmidrule(lr){4-4}
$\boldsymbol{\tau}$ & {\bf alt}$_{\textbf{120B}}$ & {\bf alt}$_{\textbf{14B}}$ & {\bf alt}$_{\textbf{120B}}$ \\
\midrule
\multirow{2}{*}{0.2} & $-2.6$ & $-0.8$ & $-3.2$ \\
& {\scriptsize[$-6.7$, $+1.3$]} & {\scriptsize[$-4.1$, $+2.5$]} & {\scriptsize[$-7.9$, $+1.5$]} \\
\multirow{2}{*}{0.4} & $-0.2$ & $+0.8$ & $\boldsymbol{-8.6}$ \\
& {\scriptsize[$-4.9$, $+5.0$]} & {\scriptsize[$-3.9$, $+5.5$]} & {\scriptsize[$-13.9$, $-3.4$]} \\
\multirow{2}{*}{0.6} & $+0.2$ & $+3.8$ & $\boldsymbol{-8.4}$ \\
& {\scriptsize[$-4.5$, $+5.2$]} & {\scriptsize[$-0.8$, $+8.7$]} & {\scriptsize[$-14.6$, $-2.2$]} \\
\midrule
No-hint & \multicolumn{2}{c}{$10.8 \rightarrow 29.7$} & $25.3 \rightarrow 49.8$ \\
\bottomrule
\end{tabular}
\caption{Pass@1 DiD (points) between hint prefixes from $h(q)$ and from alternative verified solutions at hint ratio $\tau$. Brackets report 95\% question-level bootstrap confidence intervals; bold marks intervals that exclude zero. The last row reports no-hint pass@1 of the base model and DAHS+BHA on the same questions.}
\label{tab:prefix-did}
\end{table}

\textbf{DAHS+BHA gains are not specific to the teacher hints seen in training.}~Because rollouts that receive a hint prefix continue from the verified teacher hint $h(q)$, the policy may learn to continue these specific hints rather than to solve the questions. To test this, we reveal a hint prefix on training questions either from $h(q)$ or from a verified solution to the same question that does not appear in training, written by \model{gpt-oss-120b} (alt$_{\text{120B}}$) or by the \model{Qwen3-14B} teacher (alt$_{\text{14B}}$). Because the two prefix sources may differ in difficulty, we compare their pass@1 gap before and after DAHS+BHA training and report the change as a difference-in-differences (DiD); a positive DiD means that training favored $h(q)$. No DiD in Table~\ref{tab:prefix-did} is positive with an interval that excludes zero, and on \model{Qwen3-4B-Base} the DiD is negative at all three hint ratios, whereas no-hint pass@1 on the same questions rises far beyond the base model. Likewise, the SFT-on-hints baseline, which imitates the teacher hints directly, trails DAHS+BHA at pass@1 on each AIME benchmark (Table~\ref{tab:aime-results}). Together with the absence of gradient on hint tokens (Sec.~\ref{sec:suffix-only}) and the shrinking visible prefix under BHA, these results indicate that the gains come from solving the training questions rather than from continuing their teacher hints (Appendix Sec.~\ref{app:prefix-did}).

\textbf{DAHS+BHA restores informative updates on challenging questions.}~On \model{Qwen3-1.7B-Base}, 39.7\% of training questions remain unsolved throughout vanilla DAPO training and yield no informative update, whereas under DAHS+BHA the unsolved fraction among questions that receive hints stays below 5\% (Fig.~\ref{fig:training-dynamics-dapo-bha}); hint scaffolding thus turns previously unsolved questions into informative updates. These footholds matter especially at high $k$: raising a question's success probability from $0$ to a small $\varepsilon$ adds only $\varepsilon$ to pass@1 but up to $\min(1,k\varepsilon)$ to pass@$k$.

\textbf{DAHS+BHA preserves sample-level diversity relative to DAPO.}~On \model{Qwen3-1.7B-Base}, DAHS+BHA yields 13--28\% more distinct final answers per question than DAPO on AIME24/25/26 and is more diverse under self-BLEU, length dispersion, and the Vendi score, although both RL methods remain less diverse than the base model. Token-level entropy is slightly lower under DAHS+BHA, but it measures local generation uncertainty rather than question-level coverage (Appendix Sec.~\ref{app:diversity}).

\section{Ablations}
\label{sec:ablation}

\subsection{Design of the Hint Ratio \texorpdfstring{$\tau$}{tau}}
\label{sec:ablation-tau-main}
We compare three designs for controlling hint exposure: a single global schedule, per-prompt search, and BHA's length-bucketed $\tau_b$.
A single global $\tau$ with linear annealing removes hint tokens from harder questions earlier than needed, returning them to uniformly incorrect groups; the fraction of unsolved questions then rises and evaluation metrics plateau.
Length-bucketed annealing avoids this: a bucket reduces its hint ratio only after its questions are solved often enough, so harder questions keep longer hint prefixes and continue to provide training signal.
It does not remove the risk entirely: across our \model{Qwen3-1.7B-Base} runs, 14.1\% ($B{=}50$) and 13.9\% ($B{=}200$) of the questions that the policy had solved with a hint prefix before a hint-ratio reduction remain unsolved for the rest of training, mostly in the long-hint buckets (Appendix Sec.~\ref{app:bucket-risk}).
Per-prompt Hint-Limited Search is finer-grained but shows no clear advantage in Table~\ref{tab:aime-results} at 2.7--3$\times$ BHA's online rollout overhead early in training (Appendix Sec.~\ref{sec:ablation-tau-design}); we therefore adopt length-bucketed annealing as the default.

\subsection{DAHS Ablation}
\label{sec:ablation-dahs}
Under the same BHA training recipe, replacing DAHS with non-aligned teacher hints on \model{Qwen3-1.7B-Base} lowers pass@2048 by about 6.6 points on each AIME benchmark while leaving pass@1 nearly unchanged (Appendix Table~\ref{tab:dahs-ablation}).
This gap is consistent with Fig.~\ref{fig:dahs-logprob-compare}, where DAHS hints have higher log-probability under the student policy than non-aligned hints.
A plausible mechanism, which the pass@$k$ results do not establish directly, is that DAHS hint prefixes provide continuation contexts with less policy mismatch.
These results suggest that, on \model{Qwen3-1.7B-Base}, DAHS is a key contributor to the high-$k$ gains of our method.
On \model{Llama-3.2-1B-Instruct}, the effect is less consistent across benchmarks.

\subsection{Per-Question Hint Dropout Ratio}
\label{sec:ablation-dropout}
Per-question hint dropout (Sec.~\ref{sec:bha}) balances early learning from hint prefixes against transfer to no-hint evaluation: keeping hints throughout training ($p_0=0$) hurts final no-hint performance, whereas $p_0=1$ reduces the method to DAPO and loses the early updates that hint prefixes enable.
Both $p_0=0.5$ and our default $p_0=0.75$ exceed both endpoints at pass@1 on all three AIME benchmarks; at pass@2048, $p_0=0.75$ is highest on AIME25 and AIME26 and $p_0=0.5$ on AIME24 (Appendix Sec.~\ref{app:dropout-ablation}).

\section{Related Work}

Among teacher-guided methods, one line of work optimizes the guided off-policy portion directly with supervised-style losses~\citep{liu2025uft,zhang2026onpolicyrlmeetsoffpolicy,dou2025planactionhighlevelplanningguidance}. Another incorporates guidance through off-policy weighting or correction~\citep{nath2025adaptiveguidanceacceleratesreinforcement,yan2025learningreasonoffpolicyguidance,huang2025blendingsupervisedreinforcementfinetuning}. By contrast, the closest setting to ours is suffix-only scaffolded RL, where teacher solutions, scaffolds, or hints serve only as prefix context and policy updates apply only to student-generated suffix tokens~\citep{zhang2025breadbranchedrolloutsexpert,zhang2025scaf}.

Within this suffix-only setting, BREAD (Sec.~\ref{sec:backward-algorithm}) is the closest prior work. Our work departs from BREAD in two ways. First, BREAD narrows the expert--student gap by choosing how much of a fixed expert trace to reveal, whereas DAHS changes the trace itself by synthesizing verified hints aligned with the student's continuation distribution. Second, BHA samples no-hint groups through per-question dropout and shortens scaffolds via bucketed backward annealing.

Other hint-based RLVR designs include StepHint~\citep{zhang2025stephint} and HINT~\citep{wang2025hinthelpingineffectiverollouts}; Appendix Sec.~\ref{app:extended-related-work} compares their design choices with ours.
PieceHint~\citep{fang-shi-2026-placing} withdraws hints during training, removing LLM-scored critical steps of a ground-truth solution on a fixed per-question schedule.
BHA instead shortens student-aligned hint prefixes per length bucket as bucket accuracy rises.

\section{Conclusion}
We study math RLVR through a concrete failure pattern: challenging questions can remain without informative updates for long periods, letting easier questions dominate actor updates. Distribution-Aligned Hint Synthesis (DAHS) and Backward Hint Annealing (BHA) address this by broadening continuation contexts early in training and then annealing them back to the no-hint evaluation distribution. Given verified teacher prefixes, this combination mitigates DAPO's contraction of high-$k$ problem coverage on AIME24/25/26 where it appears: on \model{Qwen3-1.7B-Base} and \model{Qwen3-4B-Base}, it improves over DAPO at both pass@1 and pass@2048, while on \model{Llama-3.2-1B-Instruct} the gains appear primarily in the high-$k$ regime without recovering the base model's full coverage. Three design choices support these gains in our setting: distribution-aligned hint design, annealing that preserves no-hint transfer, and bucket-level schedules with a practical cost-performance trade-off. These results suggest that hint scaffolding can counter distribution sharpening by broadening continuation contexts, which diversifies learning overall while restoring informative updates on challenging questions.

\section*{Limitations}
Hint-scaffolding and off-policy guidance methods rely on verified solution traces~\citep{zhang2025breadbranchedrolloutsexpert,zhang2025scaf,zhang2025stephint,wang2025hinthelpingineffectiverollouts,yan2025learningreasonoffpolicyguidance}. Our method likewise requires verified teacher hints, so teacher coverage bounds the practical learnable range. Verification applies to final answers rather than intermediate reasoning, and the DAHS ablation shows that the gains also depend on how well the student can continue the teacher hints (Sec.~\ref{sec:ablation-dahs}). We do not audit hint prefixes for final-answer exposure. Our \model{Qwen3-14B} experiment shows that training remains feasible with a smaller same-family teacher, although verified hint yield drops from 94.55\% to 87.8\% (Appendix Sec.~\ref{app:teacher-yield}). Controlled comparisons across teacher scales and model families remain future work. The evidence comes from single-seed RL runs and concentrates on Qwen3 models up to 4B parameters and on AIME, with \model{Llama-3.2-1B-Instruct} marking a boundary case. Each AIME benchmark has 30 questions; the question-level bootstrap quantifies the resulting finite-question uncertainty but not training-seed variance. We do not probe whether the base models encountered the evaluation questions during pretraining, although AIME26 postdates the release of all three base models. The hint-dropout ratio $p_0$ and the bucket count $B$ were compared on the AIME evaluation benchmarks rather than on a held-out validation set, so these ablations are exploratory. Evolving-policy diagnostics for Fig.~\ref{fig:dahs-logprob-compare} and head-to-head comparisons with other hint-based RLVR designs, whose training-data and rollout configurations differ from ours, remain future work. Appendix Sec.~\ref{app:extended-limitations} expands these points with pointers to the supporting analyses.

\bibliography{references}

\appendix
\section{Methodological Details}
\FloatBarrier
\counterwithin{figure}{section}
\counterwithin{table}{section}
\counterwithin{algorithm}{section}
\setcounter{dbltopnumber}{3}
\renewcommand{\dbltopfraction}{0.95}

\subsection{BHA Training Algorithm and Default Hyperparameters}
\label{app:training-alg}

\begin{figure*}[!t]
\small
\captionsetup[algorithm]{justification=raggedright,singlelinecheck=false}
\hrule height 0.8pt
\vspace{2pt}
\captionof{algorithm}{Backward Hint Annealing with dynamic sampling.}
\label{alg:training-alg}
\vspace{2pt}
\hrule height 0.4pt
\vspace{2pt}
\begin{algorithmic}[1]
\State \textbf{Input:} filtered training set $\mathcal{D}$ with initial verified teacher hints $\{h(q)\}$, fixed bucket map $b(q)\in\{1,\dots,B\}$, number of buckets $B$, prompts per update $N_{\mathrm{batch}}$, candidates per attempt $N_{\mathrm{cand}}$, rollouts per prompt $G$, hint-dropout probability $p_0$, initial bucket hint ratio $\tau_{b_{\mathrm{init}}}$, annealing threshold $\eta$, step size $\Delta\tau$, cutoff window $D$, token budget $L_{\max}$
\State \textbf{Initialize:} $\mathcal{H}(q) \gets [h(q)]$ for each question $q \in \mathcal{D}$
\State \hspace{\algorithmicindent} $\tau_{\texttt{b}} \gets \tau_{b_{\mathrm{init}}}$ for each bucket $\texttt{b}\in\{1,\dots,B\}$
\Loop
  \State Initialize batch $\mathcal{B}=\{\}$
  \Repeat
    \State $X \gets \textsc{Draw}(\mathcal{D}, N_{\mathrm{cand}})$
    \For{\textbf{each} $\texttt{x} \in X$}
      \State $\mathrm{uid} \gets \texttt{x.uid}$
      \State $\texttt{q} \gets \texttt{x.question}$
      \State $\texttt{b} \gets b(\texttt{q})$
      \If{$\texttt{rand()} < p_0$}
        \State $\texttt{drop}(\texttt{x}) \gets \texttt{true}$
        \State $\texttt{prefix}(\texttt{x}) \gets \emptyset$ \Comment{per-question hint dropout}
        \State $\texttt{max\_new\_tokens}(\texttt{x}) \gets L_{\max}$
      \Else
        \State $\texttt{drop}(\texttt{x}) \gets \texttt{false}$
        \State $h^\star(\texttt{q}) \sim \mathcal{H}(\texttt{q})$
        \State $\tau \gets \tau_{\texttt{b}}$
        \State $\texttt{cutoff0} \gets \texttt{round}(\tau \cdot |h^\star(\texttt{q})|)$
        \State $\texttt{cutoff} \sim \texttt{DiscreteUniform}(\{\max(0,\texttt{cutoff0}-D), \dots, \texttt{cutoff0}\})$
        \State $\texttt{prefix}(\texttt{x}) \gets \mathrm{prefix}(h^\star(\texttt{q}), \texttt{cutoff})$
        \State $\texttt{max\_new\_tokens}(\texttt{x}) \gets L_{\max} - |\texttt{prefix}(\texttt{x})|$
      \EndIf
    \EndFor
    \State \texttt{Generate} $G$ rollouts per prompt in $X$ conditioned on $\texttt{prefix}(\texttt{x})$
    \State \texttt{Score} each rollout by verifier to obtain $z_{\mathrm{uid},i}\in\{0,1\}$
    \For{\textbf{each} $\texttt{x} \in X$ with $\texttt{drop}(\texttt{x})=\texttt{true}$}
      \State $\texttt{q} \gets \texttt{x.question}$
      \For{\textbf{each} verified-correct rollout $o$ of $\texttt{x}$}
        \State Append $o$ to $\mathcal{H}(\texttt{q})$ and keep only the latest five verified hints
      \EndFor
    \EndFor
    \For{\textbf{each} $\texttt{x} \in X$}
      \State $\mathrm{uid} \gets \texttt{x.uid}$
      \State $\bar{z}_{\mathrm{uid}} \gets \frac{1}{G}\sum_{i=1}^{G} z_{\mathrm{uid},i}$ \Comment{prompt-level accuracy}
    \EndFor
    \For{\textbf{each} bucket $\texttt{b}$}
      \State $\mathcal{U}_{\texttt{b}} \gets \{\texttt{x.uid} : \texttt{x}\in X,\ \texttt{prefix}(\texttt{x})\neq\emptyset,\ b(\texttt{x.question})=\texttt{b}\}$
      \State $\bar{z}_{\texttt{b}} \gets \mathrm{mean}_{\mathrm{uid}\in \mathcal{U}_{\texttt{b}}}(\bar{z}_{\mathrm{uid}})$
      \If{$\bar{z}_{\texttt{b}} \ge \eta$}
        \State $\tau_{\texttt{b}} \gets \max(\tau_{\texttt{b}} - \Delta\tau,\, 0)$ \Comment{annealing}
      \EndIf
    \EndFor
    \State $\texttt{kept} \gets \{\mathrm{uid} : 0 < \bar{z}_{\mathrm{uid}} < 1\}$ \Comment{dynamic sampling}
    \State $\mathcal{B} \gets \texttt{concat}(\mathcal{B}, \texttt{kept rollouts})$
  \Until{$\texttt{num\_prompts}(\mathcal{B}) \ge N_{\mathrm{batch}}$}
  \State $\mathcal{B} \gets \texttt{first } N_{\mathrm{batch}} \times G \texttt{ rollouts}$
  \State \texttt{Update actor} with objective function (Eq.~\ref{eq:bha-obj})
\EndLoop
\end{algorithmic}
\vspace{2pt}
\hrule height 0.8pt
\end{figure*}

We compute $\bar{z}_b$ from the prompt-level accuracies of prompts with hints generated in the current attempt.
A single update step can include multiple refill attempts with separate $\bar{z}_b$ values.
We apply $\tau_b$ decay immediately after each attempt using that attempt's estimate.

\paragraph{Default Hyperparameters}
\label{app:default-hparams}

Appendix Table~\ref{tab:hparams} lists the default hyperparameters.
For the annealing threshold, we define
\begin{equation}
\label{eq:app-margin-def}
\begin{aligned}
&n_{\mathrm{eff}} = \left(\frac{N_{\mathrm{cand}}}{B}\right) G,
\qquad
z = 1.96,
\\[2pt]
&\kappa = z \cdot \frac{0.5}{\sqrt{n_{\mathrm{eff}}}}
\end{aligned}
\end{equation}

\begin{equation}
\label{eq:app-margin-default}
\quad \kappa \approx 0.13696,\quad \eta = 0.5 + \kappa \approx 0.63696
\end{equation}

The threshold $\eta$ is therefore derived from $n_{\mathrm{eff}}$ rather than
set as a free constant. Both
extremes of $B$ are undesirable: $B{=}1$ approaches a global schedule, whose
failure mode we analyze in Sec.~\ref{sec:ablation}---hard questions lose
hints too early---while very large $B$ yields noisy per-bucket estimates. As
$B$ grows, $n_{\mathrm{eff}}$ shrinks and
$\eta$ rises, making annealing harder to trigger from noise. With our
defaults, $n_{\mathrm{eff}}{=}51.2$ and $\eta{\approx}0.637$, safely above
chance while keeping estimates stable. For $D$, the design range is a local
jitter within $0 \le D \le \Delta\tau\cdot|h|$; with $\Delta\tau{=}0.05$ and
typical hint lengths of several hundred tokens, our default $D{=}30$ lies in
this range. The dropout ratio $p_0$ and the schedule have dedicated
ablations; the remaining hyperparameters follow DAPO.

\paragraph{Realized visible-prefix statistics}
Under these defaults on \model{Qwen3-1.7B-Base}, rollouts that receive a
prefix expose 29\% of the selected trace on average and 6.1\% over the final
10\% of training steps. These values do not measure the
post-filter composition of actor updates.

To probe sensitivity to $B$ directly, we rerun BHA with $B\in\{50,200\}$
(Table~\ref{tab:b-sensitivity}). In solved-at-least-once counts (out of 30),
the default $B{=}100$ solves 21/22/22 on AIME24/25/26, versus 18/21/19 for
$B{=}50$ and 21/22/18 for $B{=}200$: the default is never below either
extreme, with the largest gaps (three to four questions) on AIME26. We read
these gaps as directionally consistent with the operating-range
argument---both extremes are weakly worse. Because each probe changes $B$ by
only a factor of two and is a single training run without seed replicates, we
do not read them as evidence that varying $B$ has a large effect.

\begin{table}[!htb]
\centering
\footnotesize
\setlength{\tabcolsep}{2.5pt}
\begin{tabular}{@{}l ll ll ll@{}}
\toprule
& \multicolumn{2}{c}{\bf AIME24} & \multicolumn{2}{c}{\bf AIME25} & \multicolumn{2}{c}{\bf AIME26} \\
\cmidrule(lr){2-3}\cmidrule(lr){4-5}\cmidrule(lr){6-7}
{\bf Run} & {\bf p@1} & {\bf p@2048} & {\bf p@1} & {\bf p@2048} & {\bf p@1} & {\bf p@2048} \\
\midrule
$B{=}50$        & 11.7 & 60.0 & 8.3 & 70.0 & 6.3 & 63.3 \\
$B{=}100$ (default) & 13.2 & 70.0 & 8.9 & 73.3 & 7.4 & 73.3 \\
$B{=}200$       & 12.4 & 70.0 & 8.9 & 73.3 & 6.5 & 60.0 \\
\bottomrule
\end{tabular}
\caption{Sensitivity of BHA to the bucket count $B$ on
\model{Qwen3-1.7B-Base}.}
\label{tab:b-sensitivity}
\end{table}

\begin{table}[!htb]
\begin{center}
\small
\setlength{\tabcolsep}{5pt}

\begin{tabular}{@{}llp{0.55\linewidth}@{}}
\toprule
{\bf Name} & {\bf Value} & {\bf Description} \\
\midrule
$G$ & $8$ & Group size (number of rollouts per prompt). \\
$N_{\mathrm{batch}}$ & $512$ & Number of prompts per rollout step after filtering. \\
$N_{\mathrm{cand}}$ & $640$ & Number of candidate prompts per sampling attempt. \\
$p_0$ & $0.75$ & No-prefix dropout probability. \\
$L_{\max}$ & $8192$ & Per-prompt budget; we limit to $L_{\max}-|\tilde{h}|$. \\
$D$ & $30$ & Stochastic cutoff window in Eq.~\ref{eq:cutoff-draw}. \\
$B$ & $100$ & Number of length buckets used in hint annealing. \\
$\tau_{b_{\mathrm{init}}}$ & $0.9$ & Initial bucket hint ratio; set $\tau_b=\tau_{b_{\mathrm{init}}}$ per bucket. \\
$\Delta\tau$ & $0.05$ & Bucket-level annealing step size. \\
$\eta$ & $0.5 + \kappa$ & Bucket-level annealing threshold. \\
$\mathrm{lr}$ & $1\times 10^{-6}$ & Learning rate used for RL training. \\
$T_{\mathrm{warmup}}$ & $10$ & Number of learning-rate warmup steps. \\
$N_{\mathrm{update}}$ & $32$ & Number of prompts per update. (\texttt{ppo\_mini\_batch\_size}) \\
$\epsilon_{\mathrm{low}}$ & $0.2$ & Lower clipping ratio used in Eq.~\ref{eq:bha-obj}. \\
$\epsilon_{\mathrm{high}}$ & $0.28$ & Upper clipping ratio used in Eq.~\ref{eq:bha-obj}. \\
\bottomrule
\end{tabular}
\end{center}
\caption{Default hyperparameters used in the main experiments.}
\label{tab:hparams}
\end{table}

\subsection{Experimental Setup}
\label{app:exp-details}

\paragraph{Model checkpoints.}
We employ the following checkpoints: \model{Qwen/Qwen3-1.7B-Base}, \model{Qwen/Qwen3-4B-Base}, \model{Qwen/Qwen3-14B}~\citep{qwen3technicalreport}, \model{meta-llama/Llama-3.2-1B-Instruct}, \model{meta-llama/Llama-3.2-1B}, \model{meta-llama/Llama-3.2-3B}~\citep{grattafiori2024llama3herdmodels}, \model{google/gemma-3-1b-pt}~\citep{gemmateam2025gemma3}, and \model{openai/gpt-oss-120b}~\citep{openai2025gptoss120bgptoss20bmodel}.

\paragraph{Decoding settings.}
We follow the evaluation settings of \citet{yue2025does}.
We decode with temperature $0.6$ using nucleus sampling with top-$p$ $0.95$.
During evaluation, the maximum generation budget per response is $16{,}384$ tokens.

\paragraph{Implementation details.}
For each question, we first generate eight student responses with a maximum length of 4,096 tokens.
We then construct the teacher-side set $\mathcal{S}(q)$ by selecting up to the four longest solutions, which reduces the effect of overly short outputs from the base model.
We adopt \model{gpt-oss-120b} as the teacher and repeatedly draw one teacher solution at a time within a retry budget of 10, keeping the first solution that passes the verifier as the initial hint $h(q)$.
If no teacher solution passes the verifier within this budget, we exclude the question from the filtered training set.
During RL training, we retain only the latest five verified hints in the hint candidate pool $\mathcal{H}(q)$.

\paragraph{Computing infrastructure.}
We run all training and evaluation on NVIDIA H200 GPUs with 140\,GB of memory each.

\paragraph{Random seeds.}
All runs use the \texttt{verl} default random seed of $1$. We therefore report single-seed results and do not perform multi-seed replication.

\subsection{Teacher Cost and Yield}
\label{app:teacher-yield}

Table~\ref{tab:teacher-scale-yield} reports verified coverage and retry cost across two teacher settings.

\begin{table}[t]
\centering
\footnotesize
\setlength{\tabcolsep}{2.5pt}
\begin{tabular}{@{}lrrl@{}}
\toprule
\textbf{Teacher} & \textbf{Yield} & \textbf{Calls} & \textbf{RL setting} \\
\midrule
\model{gpt-oss-120b} & 94.55\% & 1.39 & Main \\
\model{Qwen3-14B} & 87.8\% & 3.20 & Paired \\
\bottomrule
\end{tabular}
\caption{Verified scaffold yield and mean teacher calls per question for the default teacher and a smaller same-family teacher.}
\label{tab:teacher-scale-yield}
\end{table}

On DAPO-Math-17k, DAHS with \model{gpt-oss-120b} retains a verified hint for 16,449 of 17,398 questions (94.55\%; 5.45\% filtered) under a 10-call worst-case budget, with 92.67\% first-attempt and 93.96\% two-attempt success. Online RL adds no extra rollouts over vanilla DAPO: over the first 25\% of training steps, DAHS+BHA needs 2.33 sampling rounds per update versus 3.27 for DAPO (Sec.~\ref{sec:ablation-tau-design}).

The smaller, same-family \model{Qwen3-14B} teacher retains verified hints for 15,269 of the 17,398 deduplicated questions (87.8\%). We retrain both DAPO and DAHS+BHA on this subset and evaluate no-hint pass@$k$ on the three AIME benchmarks (solved-at-least-once counts in Table~\ref{tab:solved-counts}, last block). DAHS+BHA matches DAPO on AIME24, solves two more questions on AIME25, and solves one fewer on AIME26. Because the two teachers yield different verified-scaffold subsets (15,269 versus 16,449 questions), this comparison cannot separate teacher hint quality from scaffold coverage. We read it only as evidence that training remains feasible with a smaller same-family teacher.

\subsection{Pass@\texorpdfstring{$k$}{k} Evaluation Protocol}
\label{app:passk-protocol}

We report pass@$k$ using the unbiased estimator of \citet{chen2021evaluatinglargelanguagemodels}, following the implementation protocol of \citet{yue2025does}.
For each evaluation question $q$, we draw $n$ responses $\{o_i\}_{i=1}^n$ and score each with the rule-based verifier of its evaluation protocol, producing binary correctness indicators $z_i\in\{0,1\}$.
We compute the number of correct responses among the $n$ samples as $c(q)=\sum_{i=1}^{n} z_i$.
We then estimate pass@$k$ over the evaluation set as
\begin{equation}
\label{eq:passk}
    \mathrm{pass@}k
    \;=\;
    \mathbb{E}_{q\sim\mathcal{Q}_{\mathrm{eval}}}
    \left[
    1 - \frac{\binom{n-c(q)}{k}}{\binom{n}{k}}
    \right].
\end{equation}
For a question with success probability $p_i$, its pass@$k$ contribution
$1-(1-p_i)^k$ approaches $1$ when $p_i>0$ and remains $0$ when $p_i=0$; at
$k{=}2048$, pass@$k$ therefore comes close to counting the questions solved at
least once. On a 30-question AIME benchmark, pass@2048 therefore moves in
increments of approximately 3.3 points per question; Appendix
Sec.~\ref{app:bootstrap} quantifies the resulting uncertainty.

\section{Ablations and Analysis}

\subsection{Extended Related Work Discussion}
\label{app:extended-related-work}

\paragraph{Curriculum and backward-style RL.}
Reverse-curriculum and backward-style methods modify the training context so that the policy first succeeds from easier subproblems or states closer to success~\citep{salimans2018learningmontezumasrevengesingle,ecoffet2021goexplorenewapproachhardexploration}.
BHA follows this intuition at the prompt level: instead of reordering examples or resetting from intermediate states, it begins with longer revealed hint prefixes and gradually shortens them, yielding a hint-annealing curriculum within the same reasoning problem.

\paragraph{Exploration, diversity, and pass@$k$ coverage.}
Recent analyses distinguish pass@1 improvements that merely sharpen the model distribution from gains that broaden the set of solvable problems under larger evaluation budgets~\citep{yue2025does}.
This perspective connects our setting to exploration work on sparse-reward tasks, where preserving diverse successful trajectories is central to continued progress~\citep{salimans2018learningmontezumasrevengesingle}.
Our goal is therefore to preserve no-hint high-$k$ coverage while improving low-$k$ accuracy.

\paragraph{Design comparison with hint-based RLVR methods.}
Table~\ref{tab:design-comparison} contrasts BREAD, StepHint, HINT, Hint-Limited Search, and DAHS+BHA along five axes: hint signal, granularity, activation, rollout allocation, and policy update.
Besides the suffix-only methods closest to our setting, it includes two recent hint-based designs, StepHint and HINT; \citet{zhang2026hintsurvey} survey the broader space of hint-based RLVR methods.

\begin{table*}[!htb]
\begin{center}
\small
\setlength{\tabcolsep}{4pt}
\begin{tabular}{@{}lp{0.15\linewidth}p{0.15\linewidth}p{0.16\linewidth}p{0.14\linewidth}p{0.14\linewidth}@{}}
\toprule
{\bf Method} & {\bf Hint signal} & {\bf Granularity} & {\bf Activation} & {\bf Rollouts} & {\bf Update} \\
\midrule
BREAD & Expert solution prefix & Per-question stop-at-first-signal & All-incorrect groups & Branched from prefix & Suffix-only \\
\addlinespace[2pt]
StepHint & Stepwise CoT prefix & Per-question, multiple hint levels & Always on & Mixed hint levels in one group & Hint + suffix, positive-only on hint \\
\addlinespace[2pt]
HINT & High-level abstract hint & Per-question & Ineffective rollouts & Re-sampled with hint & Full response (hint in prompt) \\
\addlinespace[2pt]
Hint-Limited Search & Student-aligned verified prefix (DAHS) & Per-question ratio search & Decaying global hint limit & Extra search rollouts & Suffix-only \\
\addlinespace[2pt]
DAHS+BHA (Ours) & Student-aligned verified prefix (DAHS) & Per-bucket ratio $\tau_b$ & Bucket-accuracy annealing + per-question dropout & No extra rollouts over DAPO & Suffix-only \\
\bottomrule
\end{tabular}
\caption{Design-level comparison of closely related hint-based RLVR methods, following the sample-level hint taxonomy of \citet{zhang2026hintsurvey}. Entries for StepHint and HINT follow the descriptions in their papers: for StepHint, hint-prefix tokens receive positive-only (clipped) advantages; for HINT, hints enter the user prompt and the full generated response is updated.}
\label{tab:design-comparison}
\end{center}
\end{table*}

\subsection{Hint-Limited Search Baseline}
\label{sec:ablation-perprompt-search}

\paragraph{Per-prompt search.}
For a question $q$, we run a bounded binary search over $\tau\in[0,\bar{\tau}(t)]$.
For each candidate $\tau$, we reveal a prefix $\tilde{h}$ as in Sec.~\ref{sec:hint-construct}, generate a group of $G$ rollouts, and compute the prompt-level accuracy $\bar{z}_{\mathrm{uid}}$ as in Eq.~\ref{eq:prompt-acc}.
The search continues until it finds the smallest $\tau$ that yields a non-degenerate group; otherwise, dynamic sampling filters out the instance if no candidate succeeds within the search budget.

\paragraph{Decaying hint limit.}
At training step \(t \in \{1,\dots,T\}\), we define a global hint limit
\(\bar{\tau}(t) \in [0,1]\) that decays linearly:
\begin{equation}
\label{eq:hint-limit}
\begin{aligned}
\bar{\tau}(t)&=
\begin{cases}
\tau_0\left(1-\dfrac{t-1}{T_{\mathrm{eff}}}\right), & t-1 < T_{\mathrm{eff}},\\[6pt]
0, & \text{otherwise},
\end{cases}
\\[4pt]
T_{\mathrm{eff}}&=\max\!\left(1,\left\lfloor \rho T \right\rfloor\right),
\end{aligned}
\end{equation}
\(\tau_0\) is the initial hint level.
The parameter \(\rho \in (0,1]\) sets the fraction of total training steps by which the hint limit reaches zero.

\subsection{Hint Dropout Ratio Ablation}
\label{app:dropout-ablation}

\begin{figure}[!htb]
\begin{center}
\includegraphics[width=\linewidth]{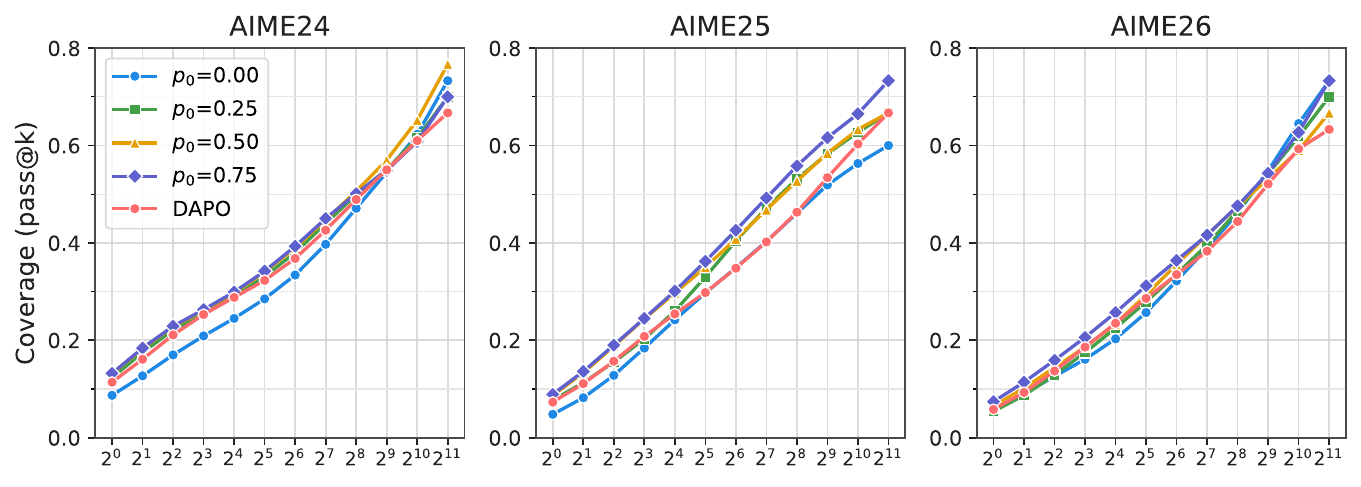}
\end{center}
\caption{Per-question hint-dropout ablation pass@$k$ curves on \model{Qwen3-1.7B-Base} for AIME24, AIME25, and AIME26.}
\label{fig:dropout-passk-curves}
\end{figure}

Table~\ref{tab:dropout-ablation} reports the endpoint metrics for the per-question hint-dropout ablation.
Fig.~\ref{fig:dropout-passk-curves} shows the full pass@$k$ curves.

\begin{table*}[!htb]
\begin{center}
\small
\setlength{\tabcolsep}{5pt}

\begin{tabular}{@{}l*{6}{c}@{}}
\toprule
& \multicolumn{2}{c}{\bf AIME24}
& \multicolumn{2}{c}{\bf AIME25}
& \multicolumn{2}{c}{\bf AIME26} \\
\cmidrule(lr){2-3}\cmidrule(lr){4-5}\cmidrule(lr){6-7}
{\bfseries\boldmath $p_0$}
& {\bf pass@1} & {\bf pass@2048}
& {\bf pass@1} & {\bf pass@2048}
& {\bf pass@1} & {\bf pass@2048} \\
\midrule

\quad 0.0 & 8.7 & 73.3 & 4.8 & 60.0 & 5.7 & \textbf{73.3} \\
\quad 0.25 & 12.3 & 70.0 & 7.7 & 66.7 & 5.4 & 70.0 \\
\quad 0.5 & 13.1 & \textbf{76.7} & 8.7 & 66.7 & 6.5 & 66.7 \\
\quad 0.75 & \textbf{13.2} & 70.0 & \textbf{8.9} & \textbf{73.3} & \textbf{7.4} & \textbf{73.3} \\
\quad 1.0 (DAPO) & 11.4 & 66.7 & 7.3 & 66.7 & 5.8 & 63.3 \\

\bottomrule
\end{tabular}
\end{center}

\caption{Ablation of the per-question hint dropout probability $p_0$ on \model{Qwen3-1.7B-Base}.}\label{tab:dropout-ablation}
\end{table*}

\subsection{Hint-Ratio Schedule Design}
\label{sec:ablation-tau-design}

This subsection reports the training-dynamics statistics behind Sec.~\ref{sec:ablation}.
Over the first 25\% of training steps, DAPO yields 63.5\% unsolved questions with 3.27 sampling rounds per update.
DAHS+BHA with $p_0=0.75$ yields 56.1\% unsolved questions with 2.33 sampling rounds per update, and DAHS+BHA with $p_0=0.5$ yields 47.9\% unsolved questions with 2.08 sampling rounds per update~(Fig.~\ref{fig:training-dynamics-dapo-bha}).
BREAD yields 8.3\% unsolved questions with 3.15 sampling rounds per update, whereas Hint-Limited Search yields 30.9\% unsolved questions with 6.19 sampling rounds per update.
BREAD attains low early degeneracy by retaining longer hints, but without explicit hint reduction its no-hint transfer stagnates later in training (Fig.~\ref{fig:no-dropout-eval-aime24}).

\subsection{DAHS Ablation}
\label{app:dahs-ablation}

\begin{table*}[!htb]
\begin{center}
\small
\setlength{\tabcolsep}{5pt}

\begin{tabular}{@{}l*{6}{c}@{}}
\toprule
& \multicolumn{2}{c}{\bf AIME24}
& \multicolumn{2}{c}{\bf AIME25}
& \multicolumn{2}{c}{\bf AIME26} \\
\cmidrule(lr){2-3}\cmidrule(lr){4-5}\cmidrule(lr){6-7}
{\bf Variant}
& {\bf pass@1} & {\bf pass@2048}
& {\bf pass@1} & {\bf pass@2048}
& {\bf pass@1} & {\bf pass@2048} \\
\midrule

\multicolumn{7}{@{}l}{\bf Qwen3-1.7B-Base} \\
\addlinespace[2pt]
\quad Non-aligned & \textbf{13.5} & 63.3 & 8.2 & 66.7 & 7.2 & 66.7 \\
\quad DAHS & 13.2 & \textbf{70.0} & \textbf{8.9} & \textbf{73.3} & \textbf{7.4} & \textbf{73.3} \\
\addlinespace[3pt]

\multicolumn{7}{@{}l}{\bf Llama-3.2-1B-Instruct} \\
\addlinespace[2pt]
\quad Non-aligned & \textbf{3.2} & 50.0 & 0.1 & \textbf{36.7} & 0.6 & 23.3 \\
\quad DAHS & 2.4 & \textbf{56.7} & 0.1 & 26.7 & 0.6 & 23.3 \\

\bottomrule
\end{tabular}
\end{center}

\caption{Ablation of Distribution-Aligned Hint Synthesis (DAHS) under the same BHA training recipe.}
\label{tab:dahs-ablation}
\end{table*}

\subsection{Impact of Flawed Reasoning Steps in Hints}
\label{app:corrupted-hint}

To test whether an incorrect reasoning step in the scaffold affects the
student's continuation, we construct a controlled corruption probe. We
decompose each verified teacher hint into a prefix, a single reasoning line
$s_k$, and a suffix, and ask the teacher (\model{gpt-oss-120b}) to corrupt
the single line into a flawed variant, yielding 1,586 valid corrupted samples
from 1,000 questions (up to three corruption attempts per question). The
student completes from three contexts: the clean prefix alone, the prefix
followed by the correct next line, and the prefix followed by the corrupted
line. Table~\ref{tab:corrupted-hint} reports pass@$k$ with 95\% bootstrap
confidence intervals for the base model and the DAHS+BHA-trained model.

Corrupted reasoning substantially lowers success, with non-overlapping
confidence intervals against prefix-only completion in all six comparisons;
for DAHS+BHA, pass@100 drops from 0.569 with the prefix alone to 0.331 with
the corrupted line. The clean next line helps directionally (0.632) with
overlapping intervals, so we make no significance claim for it. These results
indicate that the student's continuation is sensitive to the quality of the
intermediate reasoning rather than driven only by answer anchoring; however,
corruption usually also flips the final answer, so this probe does not
isolate answer-preserving corruptions. A corrupted scaffold therefore rarely
yields verified-correct rollouts and tends to produce uniformly incorrect
groups that dynamic sampling filters.

\begin{table*}[!htb]
\centering
\scriptsize
\setlength{\tabcolsep}{2.5pt}
\begin{tabular}{@{}ll lll@{}}
\toprule
{\bf Model} & {\bf Context} & {\bf pass@1} & {\bf pass@50} & {\bf pass@100} \\
\midrule
\multirow{3}{*}{Base}
& prefix            & 0.030 [0.024, 0.036] & 0.352 [0.313, 0.396] & 0.430 [0.385, 0.482] \\
& prefix + clean    & 0.050 [0.041, 0.060] & 0.428 [0.384, 0.472] & 0.499 [0.449, 0.546] \\
& prefix + corrupted& 0.008 [0.005, 0.011] & 0.153 [0.124, 0.184] & 0.219 [0.179, 0.259] \\
\addlinespace[2pt]
\multirow{3}{*}{DAHS+BHA}
& prefix            & 0.070 [0.060, 0.081] & 0.502 [0.458, 0.545] & 0.569 [0.523, 0.615] \\
& prefix + clean    & 0.091 [0.079, 0.102] & 0.570 [0.526, 0.613] & 0.632 [0.589, 0.675] \\
& prefix + corrupted& 0.017 [0.013, 0.021] & 0.247 [0.213, 0.282] & 0.331 [0.288, 0.375] \\
\bottomrule
\end{tabular}
\caption{Corruption probe: completion success from three contexts, with 95\%
bootstrap CIs. ``prefix + clean'' appends the correct next reasoning line;
``prefix + corrupted'' appends its flawed variant.}
\label{tab:corrupted-hint}
\end{table*}

\subsection{Error-Copying Audit of Teacher Solutions}
\label{app:error-copying}

DAHS conditions the teacher on student-style responses, which raises the
question of whether the teacher copies flawed student reasoning while still
reaching the verified final answer. We audit the 996 deduplicated questions
whose verified hints were used in training with a three-stage LM-judge
pipeline, using \model{gpt-5.5} (an out-of-family judge; accessed June 2026)
at temperature 0: (1) a correctness judge that takes the ground-truth answer
as given and checks the solution for internal arithmetic errors, logical
contradictions, and unjustified leaps; (2) a co-occurrence judge that decides
only whether a flagged teacher error also appears in the conditioning student
solutions, without inferring causality; and (3) an error-type classifier.

The judge flags 240/996 teacher solutions as containing a reasoning error;
78/996 (7.8\%) share an error with the conditioning student solutions, and 7
of these 78 (9.0\% of co-occurring cases; 0.7\% of all questions) contain a
concrete arithmetic error while still reaching the correct final answer.
Among the 78 co-occurring cases, 73
involve student solutions that are themselves incorrect, and unjustified
leaps dominate the error types (67/73). The conditioning pool is mostly
incorrect by construction (297 of 4,152 context solutions were judged
correct), consistent with DAHS using student responses as style templates
regardless of correctness. These results quantify a real but low-rate
error-copying channel that final-answer verification does not rule out.

\subsection{Residual Risk of Bucket-Level Annealing}
\label{app:bucket-risk}

Bucket-level annealing reduces hints for a whole bucket at once, and questions
of mixed difficulty within a bucket may therefore lose their scaffold too
early. To
quantify this residual risk, we replay each run's per-rollout hint log, which
records for every rollout the training step, the question, the bucket hint
ratio $\tau_b$, whether the rollout received a prefix, and the verifier
outcome. A question enters the analysis population if the policy solved it in
a rollout with a hint prefix at some $\tau_b$ and later received rollouts at a
strictly lower $\tau_b$. Such a question is \emph{permanently broken} if every
rollout after that solve fails, including at least one prefixed rollout at a
lower ratio, and \emph{survived} if the policy solves it again at a lower
ratio.

We report two \model{Qwen3-1.7B-Base} runs with full-training logs that use
the default setup and differ only in the bucket count. With $B{=}50$, 1,002 of the 7,130 questions in the
population are permanently broken (14.1\%) and 6,128 survive; with $B{=}200$,
1,084 of 7,784 are broken (13.9\%) and 6,700 survive. The failure rate is
therefore insensitive to the bucket count, differing by 0.2 points across a
factor-of-four change in $B$, consistent with the operating-range argument in
Sec.~\ref{app:default-hparams}.

\begin{figure}[!htb]
\begin{center}
\includegraphics[width=\linewidth]{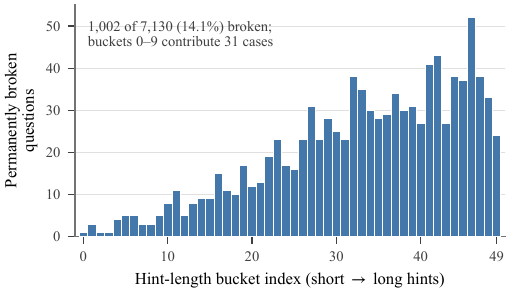}
\end{center}
\caption{Distribution of permanently broken questions across the 50
hint-length buckets of the $B{=}50$ \model{Qwen3-1.7B-Base} run
(\model{gpt-oss-120b} teacher). Broken cases concentrate in the long-hint
buckets, where annealing is slowest; the $B{=}200$ run shows the same
pattern.}
\label{fig:bucket-broken}
\end{figure}

Fig.~\ref{fig:bucket-broken} shows where the broken questions sit in the
$B{=}50$ run: buckets 0--9 contain 31 of the 1,002 broken questions while
buckets 40--49 contain 360, and the $B{=}200$ run shows the same skew. Broken
cases therefore concentrate in the long-hint buckets, which the length proxy
marks as hardest and which anneal most slowly---consistent with within-bucket
difficulty mismatch rather than overly early global annealing.

\subsection{Teacher-Hint versus Alternative-Prefix Continuation}
\label{app:prefix-did}

This section details the analysis in Table~\ref{tab:prefix-did}.

\paragraph{Questions.}
We sample 100 questions shared by the DAPO-Math-17k training subsets of the \model{Qwen3-1.7B-Base} and \model{Qwen3-4B-Base} runs, matching one question to each of the $B{=}100$ length buckets of both runs. Because DAHS conditions on each student's own responses, the verified teacher hint $h(q)$ differs between the two students.

\paragraph{Alternative solutions.}
For alt$_{\text{120B}}$, we prompt \model{gpt-oss-120b} with the question and the student's $h(q)$ and ask for a new solution whose first step uses a different idea while matching the style and length of $h(q)$, and we keep a candidate that passes the verifier. For alt$_{\text{14B}}$, we use the verified teacher hint that DAHS synthesizes for \model{Qwen3-1.7B-Base} with the \model{Qwen3-14B} teacher (Appendix Sec.~\ref{app:teacher-yield}); because this teacher is paired only with the 1.7B student, we report alt$_{\text{14B}}$ for 1.7B only. Neither alternative appears in the training runs we evaluate.

\paragraph{Hint prefixes and sampling.}
For a solution $s$ and hint ratio $\tau\in\{0.2,0.4,0.6\}$, we reveal the hint prefix $\tilde{h}=\mathrm{prefix}(s,\mathrm{round}(\tau\,|s|))$ as in Sec.~\ref{sec:hint-construct}, without the stochastic cutoff window, and prompt the student with $(q,\tilde{h})$ in the training prompt template; $\tau=0$ gives the no-hint condition shared by both sources. We evaluate the DAHS+BHA checkpoints of Table~\ref{tab:aime-results} together with their base models. For each question and condition, we draw $n{=}32$ sampled responses with temperature $0.6$, top-$p$ $0.95$, and a maximum of $8{,}192$ generated tokens.

\paragraph{Grading and statistics.}
We score each response with the same rule-based verifier as in all no-hint evaluations (Appendix Sec.~\ref{app:passk-protocol}). We grade only the generated continuation, so an answer contained in the hint prefix cannot count as correct. For each question $q$, we compute $\Delta_q$, the pass@1 with the $h(q)$ prefix minus that with the alternative prefix, and $\mathrm{DiD}_q=\Delta_q^{\mathrm{BHA}}-\Delta_q^{\mathrm{base}}$. A positive value means that BHA training increased the relative advantage of $h(q)$. Table~\ref{tab:prefix-did} reports the mean of $\mathrm{DiD}_q$ over questions with a 95\% percentile interval from a question-level bootstrap with 10,000 resamples. As in Appendix Sec.~\ref{app:bootstrap}, the interval reflects finite-question uncertainty only, and we apply no multiple-comparison correction. Table~\ref{tab:prefix-did-full} reports pass@1 for every condition.

\begin{table}[!htb]
\centering
\footnotesize
\setlength{\tabcolsep}{3.5pt}
\begin{tabular}{@{}ll*{6}{c}@{}}
\toprule
& & \multicolumn{3}{c}{\bf Base} & \multicolumn{3}{c}{\bf BHA} \\
\cmidrule(lr){3-5}\cmidrule(lr){6-8}
{\bf Model} & $\boldsymbol{\tau}$ & $\boldsymbol{h(q)}$ & {\bf alt}$_{\textbf{120B}}$ & {\bf alt}$_{\textbf{14B}}$ & $\boldsymbol{h(q)}$ & {\bf alt}$_{\textbf{120B}}$ & {\bf alt}$_{\textbf{14B}}$ \\
\midrule
\multirow{4}{*}{1.7B}
& 0   & \multicolumn{3}{c}{10.8} & \multicolumn{3}{c}{29.7} \\
& 0.2 & 12.1 & 8.7  & 13.7 & 28.1 & 27.3 & 30.6 \\
& 0.4 & 14.5 & 13.6 & 21.2 & 33.0 & 32.3 & 39.0 \\
& 0.6 & 23.1 & 17.6 & 28.2 & 45.1 & 39.4 & 46.4 \\
\addlinespace[3pt]
\multirow{4}{*}{4B}
& 0   & \multicolumn{3}{c}{25.3} & \multicolumn{3}{c}{49.8} \\
& 0.2 & 27.2 & 20.3 & -- & 52.7 & 48.9 & -- \\
& 0.4 & 32.3 & 27.8 & -- & 55.3 & 59.4 & -- \\
& 0.6 & 42.9 & 39.9 & -- & 65.3 & 70.7 & -- \\
\bottomrule
\end{tabular}
\caption{Pass@1 (\%) on the 100 training questions of Table~\ref{tab:prefix-did} for \model{Qwen3-1.7B-Base}, \model{Qwen3-4B-Base}, and their DAHS+BHA runs, with no hint ($\tau{=}0$) or a hint prefix at hint ratio $\tau$ revealed from the verified teacher hint $h(q)$ or an alternative verified solution.}
\label{tab:prefix-did-full}
\end{table}

\section{Additional Results}

\subsection{Secondary Math Benchmarks}
\label{app:secondary-benchmarks}

Table~\ref{tab:olympiadbench-math500} reports pass@1 and pass@8 on five additional math benchmarks for \model{Qwen3-1.7B-Base}, and Table~\ref{tab:secondary-4b} reports the corresponding \model{Qwen3-4B-Base} results. At 4B, DAHS+BHA matches or exceeds DAPO on pass@1 on all five benchmarks and on pass@$k$ on Olympiad, MATH500, and Minerva ($+3.8$/$+0.4$/$+3.6$), with AMC23 tied and GSM8K within $0.1$. However, both RL methods fall well below the base model's pass@128 on Olympiad, MATH500, and Minerva (for example, MATH500 $95.4 \to 72$--$73$); DAHS+BHA mitigates the contraction relative to DAPO but does not recover base coverage.

\begin{table*}[t]
\begin{center}
\small
\setlength{\tabcolsep}{1.5pt}

\begin{tabular}{@{}l*{10}{c}@{}}
\toprule
& \multicolumn{2}{c}{\bf Olympiad}
& \multicolumn{2}{c}{\bf MATH500}
& \multicolumn{2}{c}{\bf Minerva}
& \multicolumn{2}{c}{\bf AMC23}
& \multicolumn{2}{c}{\bf GSM8K} \\
\cmidrule(lr){2-3}\cmidrule(lr){4-5}\cmidrule(lr){6-7}\cmidrule(lr){8-9}\cmidrule(lr){10-11}
{\bf Method}
& {\bf pass@1} & {\bf pass@8}
& {\bf pass@1} & {\bf pass@8}
& {\bf pass@1} & {\bf pass@8}
& {\bf pass@1} & {\bf pass@8}
& {\bf pass@1} & {\bf pass@8} \\
\midrule
Base & 14.4 & 40.7 & 36.9 & \textbf{72.2} & 9.8 & 30.5 & 25.0 & 70.0 & 60.5 & 93.6 \\
DAPO & \underline{26.4} & \underline{40.9} & \underline{53.1} & 65.2 & 15.4 & 26.8 & \underline{46.9} & \textbf{77.5} & \underline{84.8} & 94.8 \\
BREAD & 25.5 & 40.7 & 52.8 & \underline{68.0} & \textbf{19.1} & \textbf{34.6} & 46.6 & 70.0 & 83.6 & \underline{95.1} \\
BHA (Ours) & \textbf{26.9} & \textbf{41.2} & \textbf{53.7} & 67.6 & \underline{17.2} & \underline{30.9} & \textbf{47.2} & \underline{72.5} & \textbf{85.0} & \textbf{95.8} \\
\bottomrule
\end{tabular}
\caption{Results on additional math benchmarks for \model{Qwen3-1.7B-Base}; pass@$k$ uses $k{=}8$ throughout.}
\label{tab:olympiadbench-math500}
\end{center}
\end{table*}

\begin{table*}[t]
\begin{center}
\small
\setlength{\tabcolsep}{1.5pt}

\begin{tabular}{@{}l*{10}{c}@{}}
\toprule
& \multicolumn{2}{c}{\bf Olympiad}
& \multicolumn{2}{c}{\bf MATH500}
& \multicolumn{2}{c}{\bf Minerva}
& \multicolumn{2}{c}{\bf AMC23}
& \multicolumn{2}{c}{\bf GSM8K} \\
\cmidrule(lr){2-3}\cmidrule(lr){4-5}\cmidrule(lr){6-7}\cmidrule(lr){8-9}\cmidrule(lr){10-11}
{\bf Method}
& {\bf pass@1} & {\bf pass@128}
& {\bf pass@1} & {\bf pass@128}
& {\bf pass@1} & {\bf pass@128}
& {\bf pass@1} & {\bf pass@8}
& {\bf pass@1} & {\bf pass@8} \\
\midrule
Base & 28.7 & \textbf{73.3} & 52.3 & \textbf{95.4} & \textbf{22.1} & \textbf{65.1} & 41.9 & 87.5 & 77.6 & \underline{97.3} \\
DAPO & \underline{34.0} & 55.9 & \underline{59.8} & 72.4 & 19.4 & 43.8 & \underline{67.8} & \textbf{90.0} & \underline{92.6} & \textbf{97.4} \\
BHA (Ours) & \textbf{35.2} & \underline{59.7} & \textbf{60.2} & \underline{72.8} & \underline{20.8} & \underline{47.4} & \textbf{70.6} & \textbf{90.0} & \textbf{93.3} & \underline{97.3} \\
\bottomrule
\end{tabular}
\caption{Secondary-benchmark results for \model{Qwen3-4B-Base} (Base/DAPO/BHA only; full baseline comparisons are reported at 1.7B). Only $n{=}8$ samples are available for AMC23 and GSM8K at 4B; those two benchmarks therefore report pass@8.}
\label{tab:secondary-4b}
\end{center}
\end{table*}

\subsection{Hint-Dropout Pass@\texorpdfstring{$k$}{k} Curves}
\label{app:dropout-passk}

\subsection{Additional Pass@\texorpdfstring{$k$}{k} Curves}
\label{app:llama-passk}

\begin{figure}[!htb]
\begin{center}
\includegraphics[width=\linewidth]{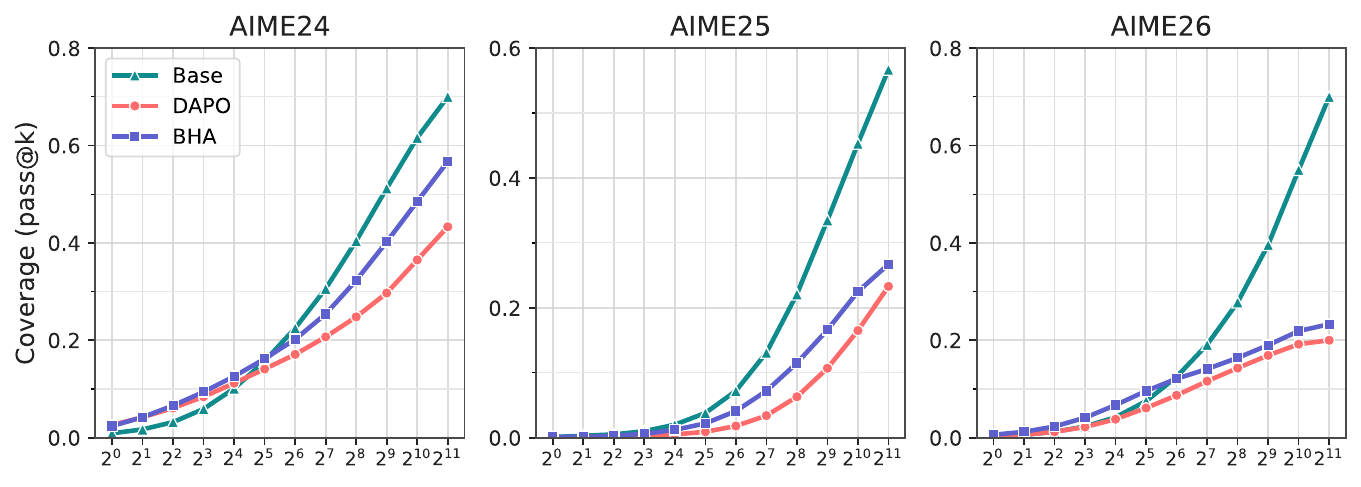}
\end{center}
\caption{Pass@$k$ curves on AIME24/25/26 ($n{=}2048$) for \model{Llama-3.2-1B-Instruct}.}
\label{fig:passk-curves-llama}
\end{figure}

Fig.~\ref{fig:passk-curves-llama} shows the pass@$k$ curves for \model{Llama-3.2-1B-Instruct}.

\subsection{Base-Model Feasibility Probe for Non-Qwen Students}
\label{app:base-boundary}

To test whether non-Qwen base checkpoints can support DAPO-style training in
our setting, we replicate the training rollout configuration (ChatML
prompt, $n=8$ rollouts per question,
temperature $1.0$, top-$p$ $1.0$, and the same DAPO reward function) on 256
questions sampled from DAPO-Math-17k, with maximum response length 8192
(2048 for \model{gemma-3-1b-pt}). The \emph{chatml} column follows this
training-faithful setting; the \emph{raw} column instead uses plain base-model
completion. Table~\ref{tab:base-boundary} reports the results.
Keepable questions under DAPO online filtering are at most 0.8\%; the
effective training signal is therefore too sparse to support stable zero-RL
training under this sampling budget. The lenient row, which searches the full output
for a boxed or ``Answer:'' string, indicates that the failure is not explained
by output formatting alone. These results delimit practical feasibility under
our setup rather than theoretical reachability. As context from a different
setup, the only sub-7B models in the
SimpleRL-Zoo study~\citep{zeng2025simplerlzooinvestigatingtamingzero} are from
the Qwen family, and its \model{Llama-3.1-8B} attains only 0.2 avg@32 on
AIME24 after zero-RL training.

\begin{table*}[!htb]
\centering
\small
\setlength{\tabcolsep}{3.5pt}
\begin{tabular}{@{}l cc cc cc@{}}
\toprule
& \multicolumn{2}{c}{\bf Llama-3.2-1B} & \multicolumn{2}{c}{\bf Llama-3.2-3B} & \multicolumn{2}{c}{\bf gemma-3-1b-pt} \\
\cmidrule(lr){2-3}\cmidrule(lr){4-5}\cmidrule(lr){6-7}
{\bf Count} & {\bf chatml} & {\bf raw} & {\bf chatml} & {\bf raw} & {\bf chatml} & {\bf raw} \\
\midrule
Verified-correct rollouts (/2048) & 1 & 0 & 1 & 2 & 0 & 0 \\
Keepable questions (/256)         & 1 & 0 & 1 & 2 & 0 & 0 \\
Lenient matches (/2048)           & 3 & 2 & 4 & 4 & 0 & 1 \\
\bottomrule
\end{tabular}
\caption{Feasibility probe for non-Qwen base checkpoints on DAPO-Math-17k
(256 questions $\times$ 8 rollouts). Keepable counts questions retained by
DAPO online filtering; lenient counts any boxed/``Answer:'' string in the
full output.}
\label{tab:base-boundary}
\end{table*}

\subsection{Evaluation Uncertainty under Question-Level Resampling}
\label{app:bootstrap}

To quantify evaluation uncertainty on the 30-question AIME benchmarks, we run
a question-level bootstrap with 10,000
resamples and report 95\% confidence intervals for pairwise differences
$\Delta$, together with one-sided bootstrap $p$-values ($p$ is the fraction of
resamples in which $\Delta \le 0$). This quantifies finite-question evaluation
uncertainty only, not training-seed variance. Tables~\ref{tab:bootstrap-qwen17b},
\ref{tab:bootstrap-qwen4b}, and \ref{tab:bootstrap-llama} report the full
comparisons for the three students. We treat the full tables as exploratory and apply no
multiple-comparison correction.
Table~\ref{tab:solved-counts} complements the bootstrap tables with exact
solved-at-least-once counts at $n{=}2048$.

\begin{table*}[!htb]
\centering
\footnotesize
\setlength{\tabcolsep}{4pt}
\begin{tabular}{@{}ll rrr r@{\,}l@{\,}l r@{\,}l@{\,}l@{}}
\toprule
{\bf Benchmark} & $k$ & {\bf Base} & {\bf DAPO} & {\bf BHA}
& \multicolumn{3}{c}{\bf $\Delta$ BHA$-$Base [95\% CI] ($p$)} & \multicolumn{3}{c}{\bf $\Delta$ BHA$-$DAPO [95\% CI] ($p$)} \\
\midrule
\multirow{4}{*}{AIME24}
& 1    & 3.34  & 11.38 & 13.25 & $+9.91$ & $[+3.62,+17.07]$ & ($<$0.001) & $+1.87$ & $[-0.32,+4.56]$ & (0.055) \\
& 64   & 33.03 & 36.80 & 39.26 & $+6.22$ & $[+0.07,+13.40]$ & (0.024)    & $+2.46$ & $[-1.07,+6.87]$ & (0.102) \\
& 1024 & 66.25 & 61.00 & 60.75 & $-5.50$ & $[-13.57,+1.77]$ & (0.925)    & $-0.25$ & $[-9.96,+9.17]$ & (0.518) \\
& 2048 & 76.65 & 66.62 & 69.98 & $-6.67$ & $[-16.67,+0.00]$ & (1.000)    & $+3.36$ & $[-10.00,+16.67]$ & (0.413) \\
\addlinespace[2pt]
\multirow{4}{*}{AIME25}
& 1    & 2.95  & 7.30  & 8.92  & $+5.97$ & $[+1.76,+11.11]$ & ($<$0.001) & $+1.62$ & $[-0.53,+4.13]$ & (0.078) \\
& 64   & 31.25 & 34.81 & 42.71 & $+11.46$ & $[+0.85,+21.64]$ & (0.017)   & $+7.90$ & $[+3.42,+13.02]$ & ($<$0.001) \\
& 1024 & 59.19 & 60.26 & 66.58 & $+7.38$ & $[-4.32,+18.91]$ & (0.110)    & $+6.32$ & $[-1.31,+15.49]$ & (0.052) \\
& 2048 & 63.44 & 66.65 & 73.42 & $+9.98$ & $[-3.33,+23.33]$ & (0.118)    & $+6.76$ & $[-6.67,+20.00]$ & (0.216) \\
\addlinespace[2pt]
\multirow{4}{*}{AIME26}
& 1    & 1.78  & 5.77  & 7.34  & $+5.56$ & $[+1.52,+10.59]$ & ($<$0.001) & $+1.57$ & $[+0.06,+3.76]$ & (0.016) \\
& 64   & 21.71 & 33.51 & 36.36 & $+14.66$ & $[+5.13,+25.33]$ & ($<$0.001)& $+2.86$ & $[-0.94,+7.03]$ & (0.075) \\
& 1024 & 62.41 & 59.19 & 62.68 & $+0.27$ & $[-11.02,+11.25]$ & (0.472)   & $+3.49$ & $[-6.57,+13.40]$ & (0.244) \\
& 2048 & 69.99 & 63.28 & 73.30 & $+3.31$ & $[-10.00,+16.67]$ & (0.417)   & $+10.01$ & $[-3.33,+23.33]$ & (0.121) \\
\bottomrule
\end{tabular}
\caption{Question-level bootstrap (10,000 resamples) for
\model{Qwen3-1.7B-Base}: pass@$k$, pairwise differences with
95\% CIs, and one-sided bootstrap $p$-values. All pass@$k$ values are
bootstrap means.}
\label{tab:bootstrap-qwen17b}
\end{table*}

\begin{table*}[!htb]
\centering
\scriptsize
\setlength{\tabcolsep}{2.5pt}
\begin{tabular}{@{}ll rrr r@{\,}l@{\,}l r@{\,}l@{\,}l r@{\,}l@{\,}l@{}}
\toprule
{\bf Benchmark} & $k$ & {\bf Base} & {\bf DAPO} & {\bf BHA}
& \multicolumn{3}{c}{\bf $\Delta$ DAPO$-$Base [95\% CI] ($p$)}
& \multicolumn{3}{c}{\bf $\Delta$ BHA$-$Base [95\% CI] ($p$)}
& \multicolumn{3}{c}{\bf $\Delta$ BHA$-$DAPO [95\% CI] ($p$)} \\
\midrule
\multirow{4}{*}{AIME24}
& 1    & 9.43  & 21.96 & 25.38 & $+12.53$ & $[+6.48,+19.65]$ & ($<$0.001) & $+15.95$ & $[+9.06,+23.81]$ & ($<$0.001) & $+3.42$ & $[+1.24,+6.06]$ & ($<$0.001) \\
& 64   & 42.84 & 60.63 & 70.86 & $+17.79$ & $[+7.76,+29.35]$ & ($<$0.001) & $+28.02$ & $[+16.19,+40.61]$ & ($<$0.001) & $+10.23$ & $[+4.27,+17.20]$ & ($<$0.001) \\
& 1024 & 74.82 & 77.68 & 81.44 & $+2.87$ & $[-7.28,+13.16]$ & (0.289)     & $+6.63$ & $[-3.00,+16.51]$ & (0.078)     & $+3.76$ & $[+0.11,+8.75]$ & ($<$0.001) \\
& 2048 & 83.30 & 79.86 & 83.23 & $-3.45$ & $[-16.67,+6.67]$ & (0.823)     & $-0.07$ & $[-10.00,+10.00]$ & (0.659)    & $+3.38$ & $[+0.00,+10.00]$ & (0.361) \\
\addlinespace[2pt]
\multirow{4}{*}{AIME25}
& 1    & 7.42  & 20.90 & 25.33 & $+13.48$ & $[+6.06,+21.93]$ & ($<$0.001) & $+17.91$ & $[+8.57,+28.07]$ & ($<$0.001) & $+4.43$ & $[+1.11,+8.44]$ & (0.003) \\
& 64   & 43.99 & 50.38 & 61.98 & $+6.39$ & $[-5.21,+18.29]$ & (0.146)     & $+17.99$ & $[+5.02,+31.33]$ & (0.002)    & $+11.61$ & $[+3.69,+21.41]$ & ($<$0.001) \\
& 1024 & 67.57 & 65.47 & 83.66 & $-2.10$ & $[-14.96,+9.88]$ & (0.627)     & $+16.09$ & $[+6.06,+27.82]$ & ($<$0.001) & $+18.19$ & $[+6.87,+31.43]$ & ($<$0.001) \\
& 2048 & 73.47 & 66.72 & 86.70 & $-6.74$ & $[-20.00,+6.67]$ & (0.911)     & $+13.23$ & $[+3.33,+26.67]$ & (0.015)    & $+19.98$ & $[+6.67,+36.67]$ & ($<$0.001) \\
\addlinespace[2pt]
\multirow{4}{*}{AIME26}
& 1    & 6.43  & 17.25 & 19.98 & $+10.82$ & $[+4.44,+18.62]$ & ($<$0.001) & $+13.55$ & $[+6.03,+22.53]$ & ($<$0.001) & $+2.73$ & $[+0.59,+5.36]$ & (0.002) \\
& 64   & 33.43 & 45.76 & 58.43 & $+12.33$ & $[+4.26,+21.68]$ & (0.001)    & $+25.01$ & $[+14.02,+36.85]$ & ($<$0.001)& $+12.67$ & $[+5.35,+20.81]$ & ($<$0.001) \\
& 1024 & 62.63 & 70.12 & 78.24 & $+7.49$ & $[-5.14,+20.73]$ & (0.125)     & $+15.60$ & $[+2.60,+30.10]$ & (0.008)    & $+8.11$ & $[-0.27,+17.84]$ & (0.029) \\
& 2048 & 69.95 & 76.64 & 80.05 & $+6.69$ & $[-10.00,+23.33]$ & (0.266)    & $+10.10$ & $[-3.33,+23.33]$ & (0.117)    & $+3.41$ & $[-6.67,+16.67]$ & (0.383) \\
\bottomrule
\end{tabular}
\caption{Question-level bootstrap (10,000 resamples) for \model{Qwen3-4B-Base}:
pass@$k$, pairwise differences with 95\% CIs, and one-sided bootstrap $p$-values.
All pass@$k$ values are bootstrap means.}
\label{tab:bootstrap-qwen4b}
\end{table*}

\begin{table*}[!htb]
\centering
\scriptsize
\setlength{\tabcolsep}{2.5pt}
\begin{tabular}{@{}ll rrr r@{\,}l@{\,}l r@{\,}l@{\,}l r@{\,}l@{\,}l@{}}
\toprule
{\bf Benchmark} & $k$ & {\bf Base} & {\bf DAPO} & {\bf BHA}
& \multicolumn{3}{c}{\bf $\Delta$ DAPO$-$Base [95\% CI] ($p$)}
& \multicolumn{3}{c}{\bf $\Delta$ BHA$-$Base [95\% CI] ($p$)}
& \multicolumn{3}{c}{\bf $\Delta$ BHA$-$DAPO [95\% CI] ($p$)} \\
\midrule
\multirow{4}{*}{AIME24}
& 1    & 0.85  & 2.64  & 2.45  & $+1.79$ & $[-0.18,+5.42]$ & (0.129)      & $+1.59$ & $[-0.04,+4.46]$ & (0.050)      & $-0.20$ & $[-1.30,+0.61]$ & (0.613) \\
& 64   & 22.49 & 17.15 & 20.22 & $-5.35$ & $[-11.01,+0.08]$ & (0.974)     & $-2.28$ & $[-5.52,+0.73]$ & (0.928)      & $+3.07$ & $[-1.26,+7.95]$ & (0.089) \\
& 1024 & 61.49 & 36.54 & 48.52 & $-24.95$ & $[-38.96,-11.40]$ & (1.000)   & $-12.97$ & $[-27.04,+1.19]$ & (0.963)    & $+11.98$ & $[+2.48,+23.28]$ & (0.005) \\
& 2048 & 69.87 & 43.43 & 56.65 & $-26.44$ & $[-43.33,-10.00]$ & (1.000)   & $-13.22$ & $[-33.33,+6.67]$ & (0.927)    & $+13.22$ & $[+3.33,+26.67]$ & (0.013) \\
\addlinespace[2pt]
\multirow{4}{*}{AIME25}
& 1    & 0.13  & 0.03  & 0.07  & $-0.10$ & $[-0.16,-0.04]$ & (1.000)      & $-0.05$ & $[-0.12,+0.01]$ & (0.941)      & $+0.05$ & $[-0.01,+0.12]$ & (0.080) \\
& 64   & 7.23  & 1.79  & 4.10  & $-5.44$ & $[-8.72,-2.56]$ & (1.000)      & $-3.13$ & $[-6.46,+0.03]$ & (0.974)      & $+2.31$ & $[-0.72,+6.27]$ & (0.088) \\
& 1024 & 45.36 & 16.50 & 22.41 & $-28.86$ & $[-42.23,-16.08]$ & (1.000)   & $-22.96$ & $[-35.16,-11.97]$ & (1.000)   & $+5.90$ & $[-6.71,+19.32]$ & (0.183) \\
& 2048 & 56.75 & 23.34 & 26.59 & $-33.41$ & $[-50.00,-16.67]$ & (1.000)   & $-30.16$ & $[-46.67,-13.33]$ & (1.000)   & $+3.25$ & $[-13.33,+20.00]$ & (0.433) \\
\addlinespace[2pt]
\multirow{4}{*}{AIME26}
& 1    & 0.30  & 0.31  & 0.62  & $+0.01$ & $[-0.18,+0.24]$ & (0.475)      & $+0.32$ & $[-0.07,+0.86]$ & (0.077)      & $+0.31$ & $[-0.00,+0.74]$ & (0.028) \\
& 64   & 12.46 & 8.71  & 12.00 & $-3.75$ & $[-9.45,+1.89]$ & (0.902)      & $-0.46$ & $[-6.53,+6.15]$ & (0.567)      & $+3.29$ & $[-0.37,+8.85]$ & (0.061) \\
& 1024 & 54.90 & 19.13 & 21.84 & $-35.77$ & $[-49.19,-22.83]$ & (1.000)   & $-33.05$ & $[-47.35,-18.96]$ & (1.000)   & $+2.71$ & $[-1.24,+9.39]$ & (0.270) \\
& 2048 & 69.98 & 19.99 & 23.31 & $-50.00$ & $[-66.67,-33.33]$ & (1.000)   & $-46.67$ & $[-63.33,-30.00]$ & (1.000)   & $+3.33$ & $[+0.00,+10.00]$ & (0.363) \\
\bottomrule
\end{tabular}
\caption{Question-level bootstrap (10,000 resamples) for
\model{Llama-3.2-1B-Instruct}: pass@$k$, pairwise differences with
95\% CIs, and one-sided bootstrap $p$-values. All pass@$k$ values are
bootstrap means. The base model retains the
broadest high-$k$ coverage; DAHS+BHA improves over DAPO (significantly on AIME24 at
$k{=}1024$/$2048$) but does not recover base coverage, consistent with the
boundary-case discussion in Sec.~\ref{sec:main-results}.}
\label{tab:bootstrap-llama}
\end{table*}

\begin{table*}[!htb]
\centering
\small
\begin{tabular}{@{}ll ccc@{}}
\toprule
{\bf Model} & {\bf Benchmark} & {\bf Base} & {\bf DAPO} & {\bf BHA} \\
\midrule
\multirow{3}{*}{Qwen3-1.7B} & AIME24 & 23 & 20 & 21 \\
                            & AIME25 & 19 & 20 & 22 \\
                            & AIME26 & 21 & 19 & 22 \\
\addlinespace[2pt]
\multirow{3}{*}{Qwen3-4B}   & AIME24 & 25 & 24 & 25 \\
                            & AIME25 & 22 & 20 & 26 \\
                            & AIME26 & 21 & 23 & 24 \\
\addlinespace[2pt]
\multirow{3}{*}{Llama-3.2-1B} & AIME24 & 21 & 13 & 17 \\
                              & AIME25 & 17 & 7  & 8  \\
                              & AIME26 & 21 & 6  & 7  \\
\addlinespace[2pt]
\multirow{3}{*}{\shortstack[l]{Qwen3-1.7B\\(14B teacher)}}
                              & AIME24 & -- & 20 & 20 \\
                              & AIME25 & -- & 19 & 21 \\
                              & AIME26 & -- & 19 & 18 \\
\bottomrule
\end{tabular}
\caption{Solved-at-least-once counts (out of 30) at $n{=}2048$ samples. On \model{Qwen3-1.7B-Base}, DAHS+BHA
exceeds DAPO on all three benchmarks and exceeds the base model on AIME25/26.
On \model{Qwen3-4B-Base}, DAHS+BHA matches the base model on AIME24 and exceeds it
on AIME25/26, while DAPO falls below the base on AIME24/25. On
\model{Llama-3.2-1B-Instruct}, DAHS+BHA improves over DAPO on all
three but remains below the base model. The last block reports the
same-family \model{Qwen3-14B}-teacher runs, discussed in
Sec.~\ref{app:teacher-yield}.}
\label{tab:solved-counts}
\end{table*}

\subsection{Solution Diversity Analysis}
\label{app:diversity}

Pass@$k$ alone is an incomplete diversity proxy, so we additionally report
sample-level diversity metrics on \model{Qwen3-1.7B-Base} over
$n{=}2048$ rollouts per question (Tables~\ref{tab:diversity}
and~\ref{tab:token-entropy}). DAHS+BHA yields 13--28\% higher unique-answer counts per
question than DAPO ($223.9{\pm}141.6$ vs $174.3{\pm}105.6$ on AIME24,
$271.9{\pm}169.5$ vs $240.6{\pm}130.7$ on AIME25, $281.0{\pm}153.4$ vs
$231.2{\pm}127.0$ on AIME26; the base model yields
$393.5$/$501.2$/$398.6$), and self-BLEU, response-length dispersion, and
Vendi scores consistently show higher sample-level diversity for DAHS+BHA than for
DAPO, while both RL methods remain less diverse than the base model.
Distinct-$n$ moves in the opposite direction: DAHS+BHA's distinct-1/2 fall below
DAPO's because pooled $n$-gram ratios decrease with response length, and DAHS+BHA
responses are substantially longer; we therefore treat distinct-$n$ as
length-confounded rather than as evidence in either direction. Token-level
entropy is slightly lower for DAHS+BHA than for DAPO. Token entropy measures local
generation uncertainty, whereas the sharpening we target concerns the
question-level solvable set, which pass@large-$k$ and solved-at-least-once
counts measure more directly.

\begin{table}[!htb]
\centering
\scriptsize
\setlength{\tabcolsep}{2.5pt}
\begin{tabular}{@{}ll ccc@{}}
\toprule
{\bf AIME} & {\bf Method} & {\bf mean entropy} & {\bf high-entropy frac.} & {\bf avg.\ tokens} \\
\midrule
\multirow{2}{*}{24} & DAPO & 0.2718 & 0.1623 $\pm$ 0.0793 & 2199 \\
                    & BHA  & 0.2532 & 0.1501 $\pm$ 0.0904 & 3528 \\
\addlinespace[2pt]
\multirow{2}{*}{25} & DAPO & 0.2705 & 0.1610 $\pm$ 0.0679 & 1610 \\
                    & BHA  & 0.2658 & 0.1574 $\pm$ 0.0763 & 2702 \\
\addlinespace[2pt]
\multirow{2}{*}{26} & DAPO & 0.2631 & 0.1564 $\pm$ 0.0692 & 2068 \\
                    & BHA  & 0.2357 & 0.1389 $\pm$ 0.0717 & 3715 \\
\bottomrule
\end{tabular}
\caption{Token-level entropy on \model{Qwen3-1.7B-Base}.}
\label{tab:token-entropy}
\end{table}

\begin{table}[!htb]
\centering
\scriptsize
\setlength{\tabcolsep}{1.5pt}
\begin{tabular}{@{}ll cccccc@{}}
\toprule
{\bf AIME} & {\bf Method} & {\bf self-BLEU$_2$\,$\downarrow$} & {\bf dist-1\,$\uparrow$} & {\bf dist-2\,$\uparrow$} & {\bf len.\ std} & {\bf Vendi\,$\uparrow$} & {\bf Vendi-emb.\,$\uparrow$} \\
\midrule
\multirow{3}{*}{24} & Base & 0.9235 & 0.00300 & 0.01219 & 2722.9 & 24.25 & 3.293 \\
                    & DAPO & 0.9781 & 0.00218 & 0.00970 & 751.8  & 11.16 & 1.478 \\
                    & BHA  & 0.9688 & 0.00143 & 0.00771 & 1587.5 & 12.06 & 1.497 \\
\addlinespace[2pt]
\multirow{3}{*}{25} & Base & 0.9164 & 0.00292 & 0.01304 & 2500.6 & 28.84 & 3.201 \\
                    & DAPO & 0.9791 & 0.00201 & 0.01016 & 489.1  & 13.17 & 1.475 \\
                    & BHA  & 0.9713 & 0.00141 & 0.00830 & 1345.5 & 14.19 & 1.484 \\
\addlinespace[2pt]
\multirow{3}{*}{26} & Base & 0.9238 & 0.00415 & 0.01615 & 2557.5 & 28.40 & 3.265 \\
                    & DAPO & 0.9752 & 0.00297 & 0.01185 & 713.0  & 14.49 & 1.489 \\
                    & BHA  & 0.9674 & 0.00178 & 0.00909 & 1641.1 & 15.81 & 1.523 \\
\bottomrule
\end{tabular}
\caption{Sample-level diversity on \model{Qwen3-1.7B-Base}
($n{=}2048$). Distinct-$n$ uses pooled $n$-grams and decreases with response
length; because DAHS+BHA responses are much longer, we treat it as
length-confounded (see text).}
\label{tab:diversity}
\end{table}

\section{Extended Limitations}
\label{app:extended-limitations}

\paragraph{Scaffold validity and teacher dependence.}
DAHS retains only questions with a verified teacher hint. Detecting misleading scaffolds that preserve the final answer while evading the verifier remains an open problem; our error-copying audit finds such concrete cases at a low but nonzero rate (7 of 996 questions; Sec.~\ref{app:error-copying}), and the corrupted-hint probe (Sec.~\ref{app:corrupted-hint}) does not cover the answer-preserving case. Beyond the same-family \model{Qwen3-14B} run (Sec.~\ref{app:teacher-yield}), we have not run RL training with a substantially smaller teacher, so we do not claim invariance to teacher quality; a lightweight reasoning-level checker over $\mathcal{H}(q)$ is future work.

\paragraph{Training and evaluation uncertainty.}
$B$ and $D$ lack dense sweeps; we provide a principled operating range and a two-point $B$ sensitivity check (Sec.~\ref{app:default-hparams}).

\paragraph{Model and benchmark scope.}
On the \model{Llama-3.2-1B-Instruct} boundary case, DAHS+BHA improves over DAPO but does not recover base coverage. We do not claim effectiveness across model families or across all math difficulty levels. We also run no head-to-head comparison with StepHint or HINT; DAPO, SFT, BREAD, and Hint-Limited Search are the closest controlled comparisons in our setting rather than a complete suite of hint-based baselines.

\paragraph{Diagnostic and curriculum limitations.}
Bucket-level annealing can misjudge within-bucket difficulty: 14\% of the questions the policy had solved with a scaffold in place become permanently broken after a hint reduction (Sec.~\ref{app:bucket-risk}).

\paragraph{Future directions.}
Because of computation constraints, we leave student models beyond 4B and additional domains to future work. Prompt-optimization frameworks such as DSPy~\citep{khattab2023dspy} and GEPA~\citep{agrawal2025gepa} are orthogonal to our training-time scaffolding and could refine the hint-synthesis prompts.

\onecolumn
\raggedbottom

\section{Qualitative Examples and Prompts}

\subsection{Prompt Templates}
\label{app:prompt-template}

We follow the instruction prompt from the DAPO-Math-17k dataset~\citep{yu2025dapo} and wrap it with each
model's default chat template.
Figure~\ref{fig:prompt-templates} shows the exact prompt formats used in our
experiments.
The \texttt{\{hint\}} field holds the hint prefix $\tilde{h}$ during training
and is empty at evaluation.

\begin{prompttemplatebox}[center,width=0.9\linewidth]{Qwen3 Prompt}
\texttt{<|im\_start|>user}\\
Solve the following math problem step by step. The last line of your
response should be of the form Answer: \$Answer (without quotes) where
\$Answer is the answer to the problem.\\
\\
\promptvar{question}\\
\\
Remember to put your answer on its own line after ``Answer:''.\texttt{<|im\_end|>}\\
\texttt{<|im\_start|>assistant}
\\
\promptvar{hint}
\end{prompttemplatebox}
\smallskip
\begin{prompttemplatebox}[center,width=0.9\linewidth]{Llama3.2 Prompt}
\texttt{<|begin\_of\_text|><|start\_header\_id|>system<|end\_header\_id|>}\\
\\
Cutting Knowledge Date: December 2023\\
Today Date: 31 Mar 2026\\
\\
\texttt{<|eot\_id|><|start\_header\_id|>user<|end\_header\_id|>}\\
\\
Solve the following math problem step by step. The last line of your
response should be of the form Answer: \$Answer (without quotes) where
\$Answer is the answer to the problem.\\
\\
\promptvar{question}\\
\\
Remember to put your answer on its own line after ``Answer:''.\texttt{<|eot\_id|><|start\_header\_id|>assistant<|end\_header\_id|>}\\
\\
\promptvar{hint}
\end{prompttemplatebox}
\captionof{figure}{Prompt templates used in our experiments. We follow the DAPO-Math-17k instruction prompt and instantiate it with each model's default chat template.}
\label{fig:prompt-templates}

\subsection{Qualitative Example}
\label{app:cot-case-analysis}

Below we present a raw chain-of-thought trace from \model{Qwen3-1.7B-Base}
trained with DAHS+BHA on one AIME25 problem, illustrating intermediate errors
and self-correction.

\begin{center}
\begin{prompttemplatebox}[width=0.9\linewidth,unbreakable,colback=questionbg,colframe=questionframe,colbacktitle=questionframe]{Question}
\textbf{(AIME25 Question No.16)}\\
Six points $A, B, C, D, E,$ and $F$ lie in a straight line in that order. Suppose that $G$ is a point not on the line and that $AC=26, BD=22, CE=31, DF=33, AF=73, CG=40,$ and $DG=30.$ Find the area of $\triangle BGE.$
\end{prompttemplatebox}
\smallskip
\begin{prompttemplatebox}[width=0.9\linewidth,colback=dahsbg,colframe=dahsframe,colbacktitle=dahsframe]{\model{Qwen3-1.7B-Base} with DAHS+BHA}
To solve the problem, let's break it down into manageable steps:

\medskip
\hrule
\medskip

\textbf{Step 1: Understand the Problem}

We have six points \( A, B, C, D, E, F \) lying in a straight line in that order. There is a point \( G \) not on this line. We are given the following distances:

\begin{itemize}
    \item \( AC = 26 \)
    \item \( BD = 22 \)
    \item \( CE = 31 \)
    \item \( DF = 33 \)
    \item \( AF = 73 \)
    \item \( CG = 40 \)
    \item \( DG = 30 \)
\end{itemize}

We need to find the area of triangle \( \triangle BGE \).

\medskip
\hrule
\medskip

\textbf{Step 2: Assign Coordinates to the Points}

Let's assign coordinates to the points on the line \( A, B, C, D, E, F \) in order along the x-axis.

Let's denote:

\begin{itemize}
    \item \( A = 0 \)
    \item \( B = b \)
    \item \( C = c \)
    \item \( D = d \)
    \item \( E = e \)
    \item \( F = f \)
\end{itemize}

From the given distances:

\begin{enumerate}
    \item \( AC = 26 \):

    \( c - 0 = 26 \) \(\Rightarrow\) \( c = 26 \).

    \item \( BD = 22 \):

    \( d - b = 22 \) \(\Rightarrow\) \( d = b + 22 \).

    \item \( CE = 31 \):

    \( e - c = 31 \) \(\Rightarrow\) \( e = c + 31 = 26 + 31 = 57 \).

    \item \( DF = 33 \):

    \( f - d = 33 \) \(\Rightarrow\) \( f = d + 33 = (b + 22) + 33 = b + 55 \).

    \item \( AF = 73 \):

    \( f - 0 = 73 \) \(\Rightarrow\) \( f = 73 \).
\end{enumerate}

From Step 3, we have \( f = b + 55 \), and from Step 5, \( f = 73 \). Therefore:

\[
b + 55 = 73 \Rightarrow b = 73 - 55 = 18.
\]

Now, find the coordinates:

\begin{itemize}
    \item \( A = 0 \)
    \item \( B = 18 \)
    \item \( C = 26 \)
    \item \( D = b + 22 = 18 + 22 = 40 \)
    \item \( E = 57 \)
    \item \( F = 73 \)
\end{itemize}

So, the coordinates are:

\begin{itemize}
    \item \( A = 0 \)
    \item \( B = 18 \)
    \item \( C = 26 \)
    \item \( D = 40 \)
    \item \( E = 57 \)
    \item \( F = 73 \)
\end{itemize}

\medskip
\hrule
\medskip

\textbf{Step 3: Determine the Coordinates of Point \( G \)}

Point \( G \) is not on the line, and we are given:

\begin{itemize}
    \item \( CG = 40 \):

    The distance from \( C \) to \( G \) is 40. Since \( C = 26 \), \( G \) lies along the line \( x = 26 + 40 = 66 \).

    \item \( DG = 30 \):

    The distance from \( D \) to \( G \) is 30. Since \( D = 40 \), \( G \) lies along the line \( x = 40 - 30 = 10 \).
\end{itemize}

\textbf{\textcolor{red}{Wait! This seems contradictory because \( G \) cannot have two different \( x \)-coordinates. Let's re-examine this.}}

From \( CG = 40 \), \( G \) is 40 units away from \( C = 26 \), so:

\[
G_x = 26 + 40 = 66.
\]

From \( DG = 30 \), \( G \) is 30 units away from \( D = 40 \), so:

\[
G_x = 40 - 30 = 10.
\]

This implies \( G \) cannot have two different \( x \)-coordinates, so there must be a misunderstanding.

\textbf{\textcolor{red}{Wait, let's re-examine the problem.}}

\textbf{\textcolor{red}{Wait! \( DG = 30 \) implies the distance from \( D = 40 \) to \( G \) is 30, so:}}

\[
G = (x, y)
\]

But since \( G \) is not on the line, it lies outside the line \( A, B, C, D, E, F \).

\textbf{\textcolor{red}{Wait, perhaps we need to find the y-coordinate.}}

We know \( G \) has coordinates \( (G_x, G_y) \).

From \( CG = 40 \), using distance formula:

\[
\sqrt{(G_x - 26)^2 + (G_y - 0)^2} = 40 \Rightarrow (G_x - 26)^2 + G_y^2 = 1600.
\]

From \( DG = 30 \):

\[
\sqrt{(G_x - 40)^2 + G_y^2} = 30 \Rightarrow (G_x - 40)^2 + G_y^2 = 900.
\]

Subtract the second equation from the first:

\[
(G_x - 26)^2 + G_y^2 - [(G_x - 40)^2 + G_y^2] = 1600 - 900.
\]

Simplify:

\[
(G_x - 26)^2 - (G_x - 40)^2 = 700.
\]

Expand both squares:

\[
(G_x^2 - 52G_x + 676) - (G_x^2 - 80G_x + 1600) = 700.
\]

Simplify:

\[
-52G_x + 676 + 80G_x - 1600 = 700.
\]

Combine like terms:

\[
28G_x - 924 = 700.
\]

Solve for \( G_x \):

\[
28G_x = 1624 \Rightarrow G_x = \frac{1624}{28} = 58.
\]

Now, substitute \( G_x = 58 \) into the second equation:

\[
(58 - 40)^2 + G_y^2 = 900 \Rightarrow 18^2 + G_y^2 = 900 \Rightarrow 324 + G_y^2 = 900.
\]

Solve for \( G_y^2 \):

\[
G_y^2 = 900 - 324 = 576 \Rightarrow G_y = \pm 24.
\]

So, \( G \) has coordinates \( (58, 24) \) or \( (58, -24) \).

For simplicity, let's take \( G = (58, 24) \).

\medskip
\hrule
\medskip

\textbf{Step 4: Find Coordinates of Points \( B, G, \) and \( E \)}

We have:

\begin{itemize}
    \item \( B = 18 \) (so \( B = (18, 0) \))
    \item \( G = (58, 24) \)
    \item \( E = 57 \) (so \( E = (57, 0) \))
\end{itemize}

\textbf{\textcolor{red}{Wait, actually, since \( B \) is on the line \( y = 0 \), and \( E \) is on the line \( x = 57 \), let's denote:}}

\begin{itemize}
    \item \( B = (18, 0) \)
    \item \( G = (58, 24) \)
    \item \( E = (57, 0) \)
\end{itemize}

\textbf{\textcolor{red}{Wait, \( E = 57 \), so \( E = (57, 0) \).}}

\medskip
\hrule
\medskip

\textbf{Step 5: Calculate the Area of \( \triangle BGE \)}

The vertices of \( \triangle BGE \) are:

\begin{itemize}
    \item \( B = (18, 0) \)
    \item \( G = (58, 24) \)
    \item \( E = (57, 0) \)
\end{itemize}

We can use the shoelace formula to find the area of the triangle:

\[
\text{Area} = \frac{1}{2} |x_1(y_2 - y_3) + x_2(y_3 - y_1) + x_3(y_1 - y_2)|
\]

Substitute the coordinates:

\[
\begin{aligned}
x_1 &= 18, & y_1 &= 0, \\
x_2 &= 58, & y_2 &= 24, \\
x_3 &= 57, & y_3 &= 0.
\end{aligned}
\]

Plug into the formula:

\[
\text{Area} = \frac{1}{2} |18(24 - 0) + 58(0 - 0) + 57(0 - 24)|
\]

Simplify:

\[
\text{Area} = \frac{1}{2} |18 \times 24 + 58 \times 0 + 57 \times (-24)|
\]

Calculate each term:

\[
18 \times 24 = 432,
\]

\[
58 \times 0 = 0,
\]

\[
57 \times (-24) = -1368.
\]

So,

\[
\text{Area} = \frac{1}{2} |432 + 0 - 1368| = \frac{1}{2} |432 - 1368| = \frac{1}{2} |-936| = \frac{1}{2} \times 936 = 468.
\]

\medskip
\hrule
\medskip

\textbf{Step 6: Conclusion}

The area of \( \triangle BGE \) is \( 468 \).

Answer: $468$
\end{prompttemplatebox}
\end{center}

\subsection{DAHS Prompt}
\label{app:dahs-prompt}

Below we present the teacher prompt that DAHS uses to synthesize the verified
teacher hint $h(q)$ from the question and the student-style responses
$\mathcal{S}(q)$.

\begin{center}
\begin{prompttemplatebox}[width=0.9\linewidth]{DAHS Prompt}
\# System Role

You are an expert mathematical editor. Your goal is to produce a single ``Golden Copy'' solution based on a set of student attempts.

\medskip
\# Objective

You will receive one \textbf{Math Problem} and a list of \textbf{Student Solutions (1 to N)}.
Your task is to produce \textbf{one fully correct solution}.

\medskip
\# Process

\begin{enumerate}
\item \textbf{Select the Base Template:} Review the provided student solutions. Choose the one that has the clearest structure and most natural flow (even if the numbers or logic are incorrect). This solution will serve as your ``Base Template.''
\item \textbf{Correct \& Refine:} Rewrite the Base Template to be mathematically perfect.
\begin{itemize}
    \item \textbf{Mimicry:} Keep the chosen student's unique writing style, formatting choices (bullet points, spacing, notation variables), and voice.
    \item \textbf{Surgical Editing:} When correcting an error, change the minimum amount of text necessary. If a number is wrong, change only the number. If a formula is wrong, swap it for the correct one but keep the surrounding sentence structure exactly as the student wrote it.
    \item \textbf{Resolution Matching:} If the student skips lines between steps, you must skip lines. If the student details every multiplication, you must do the same.
\end{itemize}
\end{enumerate}

\medskip
\# Hard Requirements

\begin{enumerate}
\item \textbf{Single Output:} Output only \textbf{one} unified, correct solution.
\item \textbf{No Meta-Commentary:} Do not mention which student solution you picked. Do not say ``fixing error'' or ``Student 2 wrote.'' Just present the final math.
\item \textbf{Rigorous Logic:} The mathematical path must be flawless.
\item \textbf{Final Answer:} End with the exact label ``Answer:'' and put the final result \textbf{on the same line immediately after} it.
\end{enumerate}

\medskip
\# Input Format

\textbf{Math Problem:}\\
\textcolor{promptaccent}{\{question\}}

\medskip
\textbf{Student Solutions:}\\
\promptvar{LIST\_OF\_STUDENT\_SOLUTIONS}

\medskip
\# Output Format

[Output only the corrected solution text here]
Answer: [FINAL\_VALUE]
\end{prompttemplatebox}
\end{center}

\subsection{Example Training Instance}
\label{app:hint-blocks}

Below we present one real training instance from our data. For the same
question, \model{gpt-oss-120b} produced the DAHS hint and the non-aligned
hint. The two hints differ in organization, level of detail, and writing
style.

\begin{center}
\begin{prompttemplatebox}[width=0.9\linewidth,unbreakable,colback=questionbg,colframe=questionframe,colbacktitle=questionframe]{Question}
The buttons $\{\times, +, \div\}$ on a calculator have their functions swapped. A button instead performs one of the other two functions; no two buttons have the same function. The calculator claims that
\[
2 + 3 \div 4 = 10
\quad \text{and} \quad
4 \times 2 \div 3 = 5.
\]
What does
\[
4 + 3 \times 2 \div 1
\]
equal on this calculator? Find the value.
\end{prompttemplatebox}
\smallskip
\parbox{\linewidth}{\emph{For readability, we show the full hints below. During training, the model receives only a revealed prefix of the selected hint, with prefix length controlled by $\tau_b$ (initialized at $\tau_{b_{\mathrm{init}}}=0.9$).}}
\smallskip
\begin{prompttemplatebox}[width=0.9\linewidth,colback=dahsbg,colframe=dahsframe,colbacktitle=dahsframe]{DAHS Hint}
Here is a complete and properly bracketed solution, with detailed explanation.

First, let us denote the actual operation performed by each button.
\begin{itemize}
    \item The button labelled ``$+$'' actually carries out some operation $op_{+}$.
    \item The button labelled ``$\times$'' actually carries out some operation $op_{\times}$.
    \item The button labelled ``$\div$'' actually carries out some operation $op_{\div}$.
\end{itemize}

Each of $op_{+}, op_{\times}, op_{\div}$ must be one of $\{+, \times, \div\}$ and they are all distinct (a permutation of the three operations).

The calculator claims
\[
2\;+\;3\;\div\;4 = 10,
\qquad
4\;\times\;2\;\div\;3 = 5.
\]

Interpreting these with the unknown mapping gives
\[
2\;op_{+}\;3\;op_{\div}\;4 = 10,
\qquad
4\;op_{\times}\;2\;op_{\div}\;3 = 5.
\]

\noindent\textbf{Usual precedence on a calculator}

\begin{itemize}
    \item Multiplication and division ($\times$ and $\div$) are performed before addition, and among themselves they are evaluated left-to-right.
    \item Addition is performed after any multiplication or division that appears.
\end{itemize}

\noindent\textbf{Finding the correct permutation}

We test the six possible permutations.
\begin{enumerate}
    \item $op_{+}=+,\,op_{\times}=\times,\,op_{\div}=\div$
    \[
    2+3\div4=2.75\neq10.
    \]

    \item $op_{+}=+,\,op_{\times}=\div,\,op_{\div}=\times$
    \[
    2+3\times4=14\neq10.
    \]

    \item $op_{+}=\times,\,op_{\times}=+,\,op_{\div}=\div$
    \[
    2\times3\div4=1.5\neq10.
    \]

    \item $op_{+}=\times,\,op_{\times}=\div,\,op_{\div}=+$
    \[
    2\times3+4=10
    \]
    and
    \[
    4\div2+3=2+3=5.
    \]
    This works.

    \item $op_{+}=\div,\,op_{\times}=+,\,op_{\div}=\times$

    This fails the first claim.

    \item $op_{+}=\div,\,op_{\times}=\times,\,op_{\div}=+$

    This also fails the first claim.
\end{enumerate}

Thus the only permutation that satisfies both given equations is
\[
\boxed{op_{+}= \times,\quad op_{\times}= \div,\quad op_{\div}= + }.
\]

In words: the ``$+$'' button actually multiplies, the ``$\times$'' button actually divides, and the ``$\div$'' button actually adds.

\noindent\textbf{Evaluating the required expression}

The expression to evaluate is
\[
4\;+\;3\;\times\;2\;\div\;1.
\]

Replacing each button by its actual operation gives
\[
4\;\times\;3\;\div\;2\;+\;1.
\]

Now apply the usual precedence:
\begin{enumerate}
    \item Perform the multiplication:
    \[
    4\times3=12.
    \]
    \item Perform the division:
    \[
    12\div2=6.
    \]
    \item Perform the addition:
    \[
    6+1=7.
    \]
\end{enumerate}

Therefore
\[
4 + 3 \times 2 \div 1 = 7
\]
on this calculator.

Answer: $7$
\end{prompttemplatebox}
\smallskip
\begin{prompttemplatebox}[width=0.9\linewidth,colback=nonalignedbg,colframe=nonalignedframe,colbacktitle=nonalignedframe]{Non-aligned Hint}
The three keys $+,\times,\div$ each now perform a different one of the three operations
addition $(+)$, multiplication $(\times)$, and division $(\div)$.
Let
\[
\begin{aligned}
f_{+} &= \text{operation performed when the ``+'' key is pressed},\\
f_{\times} &= \text{operation performed when the ``$\times$'' key is pressed},\\
f_{\div} &= \text{operation performed when the ``$\div$'' key is pressed}.
\end{aligned}
\]

Because the functions are swapped,
\[
\{f_{+}, f_{\times}, f_{\div}\} = \{+,\,\times,\,\div\},
\qquad \text{and all three are distinct.}
\]

There are $3! = 6$ possible permutations.
We test each permutation using the two statements the calculator makes.

\noindent\textbf{Testing the permutations}

\[
\begin{array}{c|c|c}
(f_{+},f_{\times},f_{\div}) & 2+3\div4 & 4\times2\div3 \\
\noalign{\vskip 2pt}
\hline
\noalign{\vskip 2pt}
(+,\,\times,\,\div) & 2+3/4=2.75 & 4\times2/3\approx2.67 \\
(+,\,\div,\,\times) & 2+3\times4=14 & 4\div2\times3=6 \\
(\times,\,+,\,\div) & (2\times3)/4=1.5 & (4\times2)/3\approx2.67 \\
(\times,\,\div,\,+) & 2\times3+4=10 & 4\div2+3=5 \\
(\div,\,+,\times) & (2/3)\times4\approx2.67 & (4/3)+2\approx3.33 \\
(\div,\times,\,+) & 2/3+4\approx4.67 & (4/2)+3=5
\end{array}
\]

Only the fourth permutation makes both given statements true:
\[
f_{+}= \times,\qquad f_{\times}= \div,\qquad f_{\div}= +.
\]

So on this calculator:
\begin{itemize}
    \item pressing ``+'' actually multiplies,
    \item pressing ``$\times$'' actually divides,
    \item pressing ``$\div$'' actually adds.
\end{itemize}

\noindent\textbf{Computing the required expression}

The expression to evaluate is
\[
4+3\times2\div1.
\]

Replacing each symbol by the operation it really performs gives
\[
4\;\underbrace{(+)}_{\times}\;3\;\underbrace{(\times)}_{\div}\;2\;\underbrace{(\div)}_{+}\;1
\;=\;(4\times3)\div2+1.
\]

Multiplication and division have the same precedence and are performed left-to-right:
\[
4\times3=12,\qquad 12\div2=6,
\]
then the final addition:
\[
6+1=7.
\]

Thus the calculator would display
\[
\boxed{7}.
\]
\end{prompttemplatebox}
\end{center}

\end{document}